%% file: TPAMI-ARCS.tex
\documentclass[10pt,journal,compsoc]{IEEEtran}
\ifCLASSOPTIONcompsoc
  \usepackage[nocompress]{cite}
\else
  \usepackage{cite}
\fi
\ifCLASSINFOpdf
\else
\fi
\usepackage{cite}
\usepackage{graphicx}
\usepackage{amsthm}
\usepackage{amsmath}
\usepackage{amsfonts}
\usepackage{amssymb}
\usepackage{enumerate} % for specifying (a), i., IV, etc 
\usepackage{mathrsfs} % for \mathscr D for distributions
\usepackage{mathtools} % :=f
\usepackage[colorlinks]{hyperref} % color 
\usepackage{multirow} % multirow
\usepackage{algorithm} % algorithmt
\usepackage[noend]{algpseudocode}
\usepackage{bbm} % indicator function
\usepackage[square,comma,numbers]{natbib} % citation 
\hypersetup{colorlinks,linkcolor={blue},citecolor={blue},urlcolor={red}}  
\usepackage{array}     % and/or
\usepackage{longtable} % and/or
\usepackage{hyperref} %link
\usepackage[small]{caption}
\usepackage[flushleft]{threeparttable} % table notes

\usepackage{lipsum}

\usepackage{colortbl}
\usepackage{arydshln} % dashed line

\usepackage{blkarray, bigstrut}
\usepackage{xparse}

\usepackage{setspace}

\usepackage{pgfplots}
\usetikzlibrary{arrows.meta}

\input{defs}

\begin{document}
\title{Optimizing Regularized Cholesky Score for Order-Based Learning of Bayesian Networks
\thanks{Department of Statistics, University of California, Los Angeles. 
This work was supported by NSF grant IIS-1546098. 
Email: yeqiaoling@g.ucla.edu, aaamini@stat.ucla.edu, zhou@stat.ucla.edu.}
}

\author{Qiaoling Ye, Arash A. Amini, and Qing Zhou}

\IEEEtitleabstractindextext{%
\begin{abstract}
	Bayesian networks are a class of popular graphical models that encode causal and conditional independence relations among variables by directed acyclic graphs (DAGs). 
	We propose a novel structure learning method, annealing on regularized Cholesky score (ARCS), to search over topological sorts, or permutations of nodes, for a high-scoring Bayesian network. Our scoring function is derived from regularizing Gaussian DAG likelihood, and its optimization gives an alternative  formulation of the sparse Cholesky factorization problem from a statistical viewpoint, which is of independent interest. 
	We combine global simulated annealing over permutations with a fast proximal gradient algorithm, operating on triangular matrices of edge coefficients, to compute the score of any permutation. Combined, the two approaches allow us to quickly and effectively search over the space of DAGs without the need to verify the acyclicity constraint or to enumerate possible parent sets given a candidate topological sort. The annealing aspect of the optimization is able to consistently improve the accuracy of DAGs learned by local search algorithms. 
	In addition, we develop several techniques to facilitate the structure learning, including pre-annealing data-driven tuning parameter selection and post-annealing constraint-based structure refinement. 
	Through extensive numerical comparisons, we show that ARCS achieves substantial improvements over existing methods, demonstrating its great potential to learn Bayesian networks from both observational and experimental data.
\end{abstract}

}

\maketitle

\IEEEdisplaynontitleabstractindextext

% \fi
\IEEEpeerreviewmaketitle

\IEEEraisesectionheading{\section{Introduction}\label{sec:introduction}}


\IEEEPARstart{B}{ayesian} networks (BNs) are a class of graphical models, whose structure is represented by a directed acyclic graph (DAG). They are commonly used to model causal networks and conditional independence relations among random variables. The past decades have seen many successful applications of Bayesian networks in computational biology, medical science, document classification, image processing, etc. 
	As the relationships among variables in a BN are encoded in the underlying graph, various approaches have been put forward to estimate DAG structures from data. Constraint-based approaches, such as the PC algorithm~\citep{pc_91}, perform a set of conditional independence tests to detect the existence of edges. In score-based approaches, a network structure is identified by optimizing a score function \cite{HGC95, S93}. There are also hybrid approaches, such as the max-min hill-climbing algorithm~\citep{TEA06}, which use a constraint-based method to prune the search space, followed by a search for a high-scoring network structure.

	The score-based search has been applied to three different search spaces: the DAG space \citep{HGC95, GMP11}, the equivalence classes \citep{C02, HGC95} and the ordering space (or the space of topological sorts) \citep{Larranaga1996, TK05}. 
	In this paper, we focus on the order-based search, which has two major advantages. First, the existence of an ordering among nodes guarantees a graph structure that satisfies the acyclicity constraint. Second, the space of orderings is significantly smaller than the space of DAGs or of the equivalence classes. 
	Consequently, several lines of research have developed efficient order-based methods for DAG learning. Some methods adopt a greedy search in conjunction with various operators that propose moves in the ordering space \citep{TK05,ADJ11, SCC15, SCZ17}. A greedy search, however, may easily be trapped in a local minimum, and thus different techniques were proposed to perform a more global search \citep{SM06, Larranaga1996, BC13,LB17}. In particular, stochastic optimization, such as the genetic algorithm~\citep{Larranaga1996, CPV18, SCZ17} and Markov chain Monte Carlo \citep{FK03, EW08}, has been advocated as a promising way to perform global search over the ordering space.

	In spite of these methodological and algorithmic advances, there are a few difficulties in score-based learning of topological sorts for DAGs. First, the score of an ordering is usually defined by the score of the optimal DAG compatible with the ordering. The computational complexity of finding the optimal DAG given an ordering,  typically by enumerating all possible parent sets for each node \citep{CH92}, can be as high as $O(p^{k+1})$ for $p$ nodes and a prespecified maximum indegree of $k$. Such computation is needed for every ordering evaluated by a search algorithm, which becomes prohibitive when $k$ is large. Second, although the ordering space is smaller than the graph space, optimization over orderings is still a hard combinatorial problem due to the NP-hard nature of structure learning of BNs \citep{Chi96}. It is not surprising that the performance of the above global optimization algorithms degrades severely for large graphs.

	Motivated by these challenges, we develop a new order-based method for learning Gaussian DAGs by optimizing a regularized likelihood score. Representing an ordering by the corresponding permutation matrix $P$, the weighted adjacency matrix of a Gaussian DAG can be coded into a lower triangular matrix $L$. We add a concave penalty function to the likelihood to encourage sparsity in $L$, and thus achieve the goal of structure learning. Instead of a prespecified maximum indegree, which is \textit{ad hoc} in nature, we provide a principled data-driven way to determine the tuning parameters for the penalty function. Finding the optimal DAG given $P$ is then reduced to $p$ decoupled penalized regression problems, which are solved by proximal gradient, an efficient first-order method, without enumerating possible parent sets for any node. Searching over $P$ is done by simulated annealing (SA). Other than using SA solely as a global optimization algorithm, we may also incorporate informative initial orderings, learned from an existing local algorithm, by setting a low starting temperature. 
	Our numerical results demonstrate that this strategy substantially improves the accuracy of an estimated DAG. We note an interesting connection between our formulation and the sparse Cholesky factorization problem, and thus name our scoring function the \emph{regularized Cholesky score} of orderings or permutations.

	Regularizing likelihood with a continuous penalty function has been shown to be effective in learning Gaussian DAGs \citep{FZ13,AZ15}. These methods optimize a regularized likelihood score over the DAG space by blockwise coordinate descent, which is a local search algorithm in nature and thus likely to be trapped in a suboptimal structure. Using DAGs learned by these methods to generate initial orderings, our method can significantly improve the accuracy in structure learning. This highlights the advantages of combining local and global searches over the ordering space under an annealing framework.
	More recently, Champion et al. \cite{CPV18} developed a genetic algorithm that optimizes an $\ell_1$-penalized likelihood over a triangular coefficient matrix and a permutation to learn Gaussian BNs. However, the $\ell_1$ penalty tends to introduce more bias in estimation than a concave penalty, thus producing less accurate DAGs. The authors did not provide a principled method to select the tuning parameter for the $\ell_1$ penalty. Given a permutation, they optimize the network structure by an adaption of the least angle regression \citep{EHJT04}, which is closely related to the Lasso. In contrast, we use a more general and effective first-order method, the proximal gradient algorithm, which is applicable to many regularizers, including the $\ell_1$ and concave penalties. As shown by our numerical experiments, our method substantially outperforms their genetic algorithm.

	The paper is organized as follows. Section~\ref{sec:background} covers some background on Gaussian BNs  and the role of permutations in identifying the underlying DAGs. We introduce the regularized Cholesky loss and set up the global optimization problem for BN learning in Section~\ref{sec:rc-score}. In Section \ref{sec:optim}, we develop the annealing on regularized Cholesky score (ARCS) algorithm which combines global annealing to search over the permutation space and a proximal gradient algorithm to optimize the network structure given an ordering. We also propose a constraint-based approach to prune the estimated network structure after annealing process and a data-driven model selection technique to choose tuning parameters for the penalty function. Section \ref{sec:res} consists of exhaustive numerical experiments, where we compare our method to existing ones using both observational and experimental data. We conclude  with a discussion in Section~\ref{sec:discussion}. All proofs are relegated to the Appendix.

%%%%%%%%%%%%%%%%%%%%%%%%%%%%%%%%%%%%%%%

\section{Background}\label{sec:background}

We start with some background on Bayesian networks. A {\em Bayesian network} for a set of variables $\{X_1, \ldots, X_p\}$ consists of 1) a directed acyclic graph $\calG$ that encodes a set of conditional independence assertions among the variables, and 2) a set of local probability distributions associated with each variable. It can be considered as a recipe for factorizing a joint distribution of $\{X_1, \ldots, X_p\}$ with probability density
\begin{equation}\label{eq:bn_def}
	p(x_1, \ldots, x_p) = \prod_{j=1}^p p(x_j \mid \Pi_j^\calG=pa_j), 
\end{equation}
where $\Pi_j^\calG \subseteq \{X_1, \ldots, X_p\} \setminus \{X_j\} $ is the parent set of variable $X_j$ in $\calG$ and $pa_j$ its value. 
The DAG $\calG$ is denoted by $\calG = (V, E)$, where $V = \{1, \ldots, p\}$ is the vertex set corresponding to the set of random variables and $E = \{(i,j): i \in \Pi_j^\calG\} \subseteq V \times V$ is the edge set. We use variable $X_j$ and node $j$ exchangeably throughout the paper. DAGs contain no directed cycles, making the joint distribution in~\eqref{eq:bn_def} well-defined. 

\subsection{Gaussian Bayesian networks} 

In this paper, we focus on Gaussian BNs that can be equivalently represented by a set of linear structural equation models (SEMs): 
\begin{align} \label{eq:sems}
	X_j = \sum_{i \in \Pi_j^\calG}\beta_{ij}^0 X_i + \eps_j, \quad j = 1, \ldots, p,
\end{align}
where $\eps_j \sim \calN(0, (\omega_j^0)^2)$ are mutually independent and independent of $\{X_i: i \in \Pi_j^\calG\}$. 
Defining $B_0 := (\beta_{ij}^0) \in \R^{p \times p}$, $\eps := (\eps_1, \ldots, \eps_p) ^\top\in \R^p$, and $X := (X_1, \ldots, X_p)^\top \in \R^p$, we rewrite~\eqref{eq:sems} as 
\begin{align} \label{eq:sems:vec}
	X=B_0^\top X + \varepsilon.
\end{align}
The model has two parameters: 1) $B_0$ as a coefficient matrix, sometimes called the weighted adjacency matrix, where ${\beta_{ij}^0}$ specifies a weight associated with the edge $i \to j$ and ${\beta_{ij}^0}=0$ for $i \notin \Pi_j^\calG$, and 2) $\Omega_0 := \diag((\omega_j^0)^2)$ as a noise variance matrix. The SEMs in~\eqref{eq:sems} define a joint Gaussian distribution, $X \sim \calN(0, \Sigma_0),$ where $\Sigma_0$ is positive definite and given by
\begin{equation}\label{eq:sigma_0}
	\Sigma_0^{-1} = (I-B_0) \Omega_0^{-1} (I-B_0)^{\top}.
\end{equation}

\subsection{Acyclicity and permutations}

\begin{figure}[t]
	\centering
	\includegraphics[width=.47\textwidth] {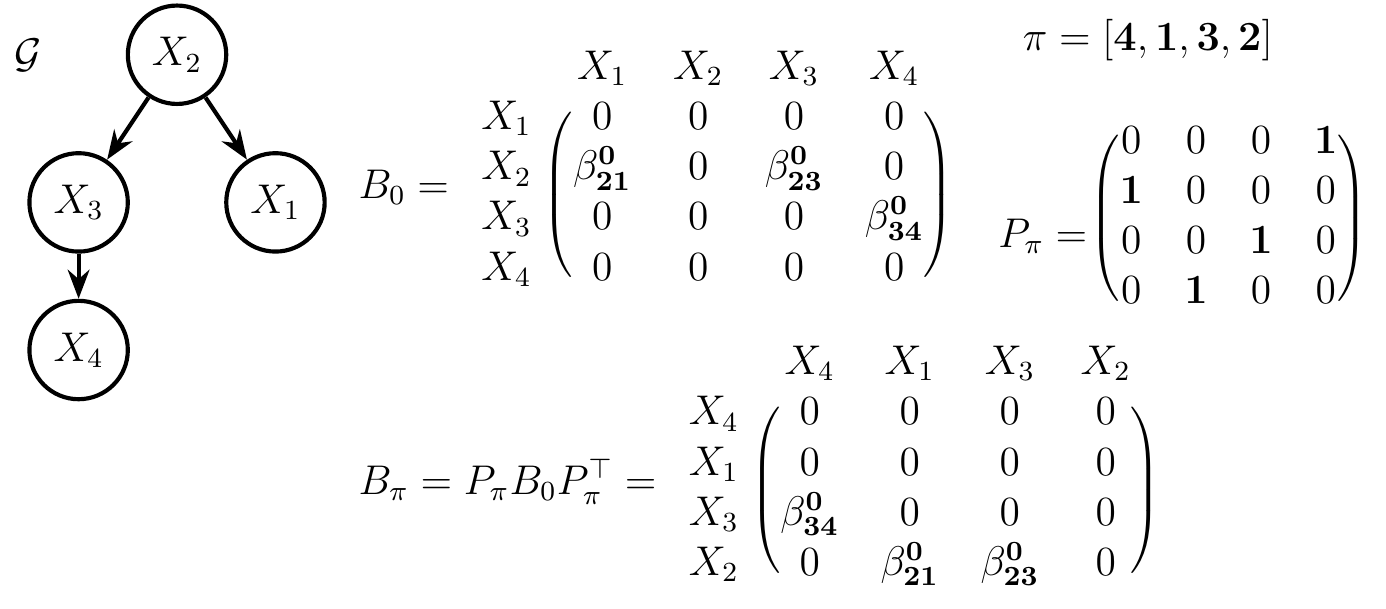}
	\caption{An example DAG $\calG$, its coefficient matrix $B_0$, and a permutation $\pi$. $B_\pi$ permutes columns and rows of $B_0$ and is strictly lower triangular.}\label{fig:topo_sort_exp}
\end{figure}

The support of $B_0$ in \eqref{eq:sems:vec} defines the structure of $\calG$, and thus it must satisfy the acyclicity constraint so that $\calG$ is indeed a DAG. To facilitate the development of our likelihood score for orderings, we express the acyclicity constraint on $B_0$ via permutation matrices. Let $\{e_1,\dots,e_p\}$ be the canonical basis of $\R^p$. To each permutation $\pi$ on the set $[p] := \{1, \ldots, p\}$, we associate a permutation matrix $P_\pi$ whose  $i^{\text{th}}$ row is $e_{\pi(i)}^\top$. For a vector $v = (v_1, \ldots, v_p)^\top$, we have 
\begin{align}\label{eq:Ppi:def}
	P_\pi v = v_\pi  = (v_{\pi(1)}, \ldots, v_{\pi(p)})^\top,
\end{align}
that is, $P_\pi$ permutes the entries of $v$ according to $\pi$. Since $P_\pi^\top P_\pi = I_p$, we can rewrite~\eqref{eq:sems:vec} as
\begin{align*}
	P_\pi X & =  B_\pi^\top P_\pi X + P_\pi \eps, \quad  
\end{align*}
where $B_\pi := P_\pi B_0 P_\pi^\top$ is obtained by permuting the rows and columns of $B_0$ simultaneously according to $\pi$. Then, $B_\pi$ will be a strictly lower triangular matrix if and only if $\pi$ is the reversal of a topological sort of $\calG$, i.e., $i\prec j$ in $\pi$ for $j\in \Pi_i^\calG$. See Figure~\ref{fig:topo_sort_exp} for an illustration. Under this reparameterization, we translate the acyclicity constraint on $B_0$ into the constraint that $B_{\pi}$ must be strictly lower triangular for some permutation $\pi$. Define $\Omega_\pi := P_\pi \Omega_0 P^\top_\pi$. Equivalently, one may think of the node $\pi(i)$ relabeled as $i$ in $B_\pi$ and $\Omega_\pi$.

For simplicity, we drop the subscript $\pi$ from  $P_\pi$, $B_\pi$ and $\Omega_\pi$ if no confusion arises. Therefore, throughout the paper, $P$ defines a permutation $\pi$, $B$ and $\Omega$ label nodes according to $\pi$ and we write the permuted SEM as
\begin{align} \label{eq:sem:per}
	P X &= B^\top P X + P \eps.
\end{align}
Denote by $\text{cov}(X)$ the covariance matrix of $X$. Then we have $\Sigma := \text{cov}(PX) =  P\Sigma_0 P^\top$, obtained by permuting the rows and columns of $\Sigma_0$~\eqref{eq:sigma_0} according to $P$.

%%%%%%%%%%%%%%%%%%%%%%%%%%%%%%%%%%%%%%%

\section{Regularized likelihood score}\label{sec:rc-score}

In this section, we formulate the objective function to estimate BN structure given data from the Gaussian SEM~\eqref{eq:sems}.

\subsection{Cholesky loss}

Let $\Xb := [\Xb_1, \ldots, \Xb_p] \in \R^{n \times p}$ be a data matrix where each row is an i.i.d. observation from~\eqref{eq:sems}.  According to~\eqref{eq:sem:per}, we obtain a similar SEM on the data matrix: 
\begin{equation} \label{eq:linear_xp}
	\Xb P^\top = \Xb P^\top B + {\bf E} P^\top,
\end{equation}
where each row of ${\bf E} \in \R^{n \times p}$ is an i.i.d. error vector from $\calN(0, \Omega_0)$. In~\eqref{eq:linear_xp}, $\Xb P^\top$  and ${\bf E}P^\top$ are $\Xb$ and ${\bf E}$ with columns permuted according to $P = P_\pi$. It then follows that each row of $\Xb P^\top$ is an i.i.d. observation from $\calN(0,\Sigma)$ with $\Sigma^{-1}=(I - B) \Omega^{-1}  (I - B)^\top$, and thus
the negative log-likelihood of~\eqref{eq:linear_xp} is 
\begin{align}
&\ell( B,\Omega, P \given \Xb) \nonumber\\
&= \f 1 2 \tr\left[ P \Xb^\top \Xb P^\top (I - B) \Omega^{-1}  (I - B)^\top\right] + \f n 2 \log|\Omega|. \label{eq:neglikobs}
\end{align}
Recall that $B$ and $\Omega = \diag((\omega_j)^2)$ are defined by permuting the rows and columns of $B_0$ and $\Omega_0$ by the permutation matrix $P$. In particular, $B$ is strictly lower triangular and we write its columns as $\beta_j\in\R^p$.

Denote by $L := (I-B)\Omega^{-\f 1 2}$ a weighted coefficient matrix, where each column $L_j = (e_j - \beta_j) / {\omega_j}$ is a weighted coefficient vector for node $\pi(j)$. We define what we call the \emph{Choleskly loss} function
\begin{align} \label{eq:chol:loss}
	\lchol(L; A) := \f 1 2\tr\big( A L L^\top\big)  -\log |L|,
\end{align}
where $|L|$ denotes the determinant of $L$. Noting that $|L| = |(I-B)\Omega^{-\f 1 2}| = |\Omega|^{-\f 1 2 }$ and denoting by $\Sigh := \f 1 n \Xb^\top \Xb$ the sample covariance matrix, one can re-parametrize the negative log-likelihood \eqref{eq:neglikobs} with $L$ and $P$ and connect it to the Cholesky loss:
\begin{lemma}
	The negative log-likelihood~\eqref{eq:neglikobs} for observational data can be re-parametrized as
	\begin{align}
		\ell( L, P) &= n \cdot  \lchol(L; P \Sigh P^\top) \label{eq:LP:loss} \nonumber \\
		&= \f n 2\tr\big(P \Sigh P^\top L L^\top\big)  -n\log |L|, 
	\end{align}
where $L = (I-B)\Omega^{-\f 1 2}$ is a lower triangular matrix and $P$ is a permutation matrix.
\end{lemma}

The reason for naming~\eqref{eq:chol:loss} the Cholesky loss is that it provides an interesting variational characterization of the Cholesky factor of the inverse of a matrix as the following proposition shows. Let $\calL_p$ be the set of $p \times p$ lower triangular matrices with positive diagonal entries, and for any positive definite matrix $M$, let $\calC(M)$ be its unique Cholesky factor, i.e., the unique lower triangular matrix $L$ with positive diagonal entries such that $M = LL^\top$.
\begin{prop} \label{prop:cf}
	For any positive definite matrix $A \in \R^{p \times p}$, we have
	\begin{align*}
		\arg\min_{L \,\in\, \calL_p} \lchol(L; A) = \left\{ \calC(A^{-1})\right\}
	\end{align*}
	with optimal value 
	\[ \lchol^*(A) := \lchol(\calC(A^{-1}); A) = \f 1 2 \left(p + \log|A|\right).\] 
	Consequently, $\lchol^*(A) = \lchol^*(PAP^\top)$ for any permutation  matrix $P$.
\end{prop}

The proof is given in Appendix~\ref{proof:prop:cf}. Proposition~\ref{prop:cf} states that $ \calC(A^{-1})$ is the unique minimizer of $\Lc_{\text{chol}}(\,\cdot\,;A)$. Now consider finding the maximum likelihood DAG for a fixed permutation $P$, which corresponds to minimizing $L \mapsto \ell(L;P)$ given by~\eqref{eq:LP:loss}. Let $\ell^*(P)$ be the optimal value of this problem, i.e., 
	\[\ell^*(P) := \min_{L\,\in\, \calL_p} \ell(L;P). \]
Then, Proposition~\ref{prop:cf} implies
\begin{align} \label{eq:ell:opt}
	\ell^*(P) = n \cdot \lchol^*(P \Sigh P^T) = n \cdot \lchol^*(\Sigh),
\end{align}
showing that $\ell^*$ is invariant to permutations, hence maximum likelihood estimation does not favor any particular ordering. In other words, all the maximum likelihood DAGs corresponding to different permutations give the same Gaussian likelihood.

\subsection{Sparse regularization}

To break the permutation equivalence of the maximum likelihood \eqref{eq:ell:opt}, we add a regularizer to the Cholesky loss to favor sparse DAGs. Under faithfulness~\citep{SGS01_causal}, the true DAG $\calG$ in \eqref{eq:sems} and its equivalence class are the sparsest among all DAGs that can parameterize the joint distribution $\calN(0, \Sigma_0)$. To start, let us point out some connections to the well-known ``sparse Cholesky factorization'' problem from linear algebra.

According to Proposition \ref{prop:cf}, the minimizer of $\ell(L;P)$ over $L$ is the Cholesky factor of $(P\widehat\Sigma P^\top)^{-1}=P\widehat\Sigma^{-1} P^\top$. For a sparse $\Sigh^{-1}$, it is well-known that the choice of $P$ greatly affects the sparsity of the resulting Cholesky factor. Heuristic approaches have been developed in numerical linear algebra to find a permutation that leads to a sparse factorization by trying to minimize the so-called ``fill-in''. An example is the maximum cardinality algorithm~\citep{VA14}. 

From a statistical perspective, however, $\Sigh^{-1}$ is, in general, not sparse (due to noise) even if the inverse of population covariance matrix $\Sigma = \ex [\Sigh]$ is so. In such cases, one can first estimate a sparse precision matrix and then use the sparse estimate as the input to the sparse Cholesky factorization problem. We take a more direct alternative approach by adding a sparsity-measuring penalty to the Cholesky loss.

Let $\rho_\theta: \R \mapsto [0, \infty)$ be a nonnegative and nondecreasing regularizer with some tuning parameter(s)~$\theta$. We consider the following penalized loss function:
\begin{align} \label{eq:fLP:def}
	f_\theta(L;P)  := ~ n \cdot \lchol(L; P \Sigh P^\top) +  \sum_{i>j} \rho_\theta(L_{ij}), 
\end{align}
where the penalty is only applied to the off-diagonal entries of a lower triangular matrix $L$. The loss depends on the regularization parameter $\theta$, and for simplicity we write $f_\theta(L;P)$ as $f(L;P)$. In this paper, we focus on the class of regularizers called the minimax concave penalty (MCP)~\citep{Z10_MCP}
which includes $\ell_1$ and $\ell_0$ as extreme cases; see~\eqref{eq:mcp} below. MCP is a sparsity-favoring penalty and adding it breaks the symmetry of the Cholesky loss w.r.t. permutations as in~\eqref{eq:ell:opt}. As a result, the permutations leading to sparser lower-triangular factors $L$ will have  smaller loss values $f(L;P)$.

Let $\calP_p$ be the set of $p \times p$ permutation matrices. Given $P\in\calP_p$, the minimizer of $f(L;P)$ over $L$ is a sparse DAG $\calG(P)$ with a score $f(P)$ defined as
\begin{align} \label{eq:fp}
	f(P) &:= \min_{L\in\calL_p} f(L;P). 
\end{align}
We can minimize permutation score $f(P)$ over $\calP_p$ to obtain an estimated ordering. The overall sparse BN learning problem is then
\begin{align}
\min_{P\in\calP_p} f(P) 
=  \min_{P\in\calP_p} \min_{L\in\calL_p} &\Big\{ \f n 2  \tr\big( P\widehat\Sigma P^{\top}L L^\top\big)  \nonumber\\
& \quad - n \log|L| + \sum_{i>j}\rho_\theta(L_{ij}) \Big\}.\label{eq:min_fp}
\end{align}
In Section~\ref{sec:optim}, we discuss our approach to solve this problem by optimizing over $(P,L)$. It is worth noting that problem~\eqref{eq:min_fp} can be considered both as 1) a penalized maximum likelihood BN estimator in the Gaussian case, and 2) a variational formulation of the sparse Cholesky factorization problem when the input matrix $\Sigh$ is noisy (hence its inverse usually not sparse). According to the second interpretation, we call $f(L;P)$ in~\eqref{eq:fLP:def} the \emph{regularized Cholesky loss} function and $f(P)$ in~\eqref{eq:fp} the \emph{regularized Cholesky (RC) score} of a permutation $P$.

\begin{figure}[t]
	\centering
	\includegraphics[width=.35\textwidth] {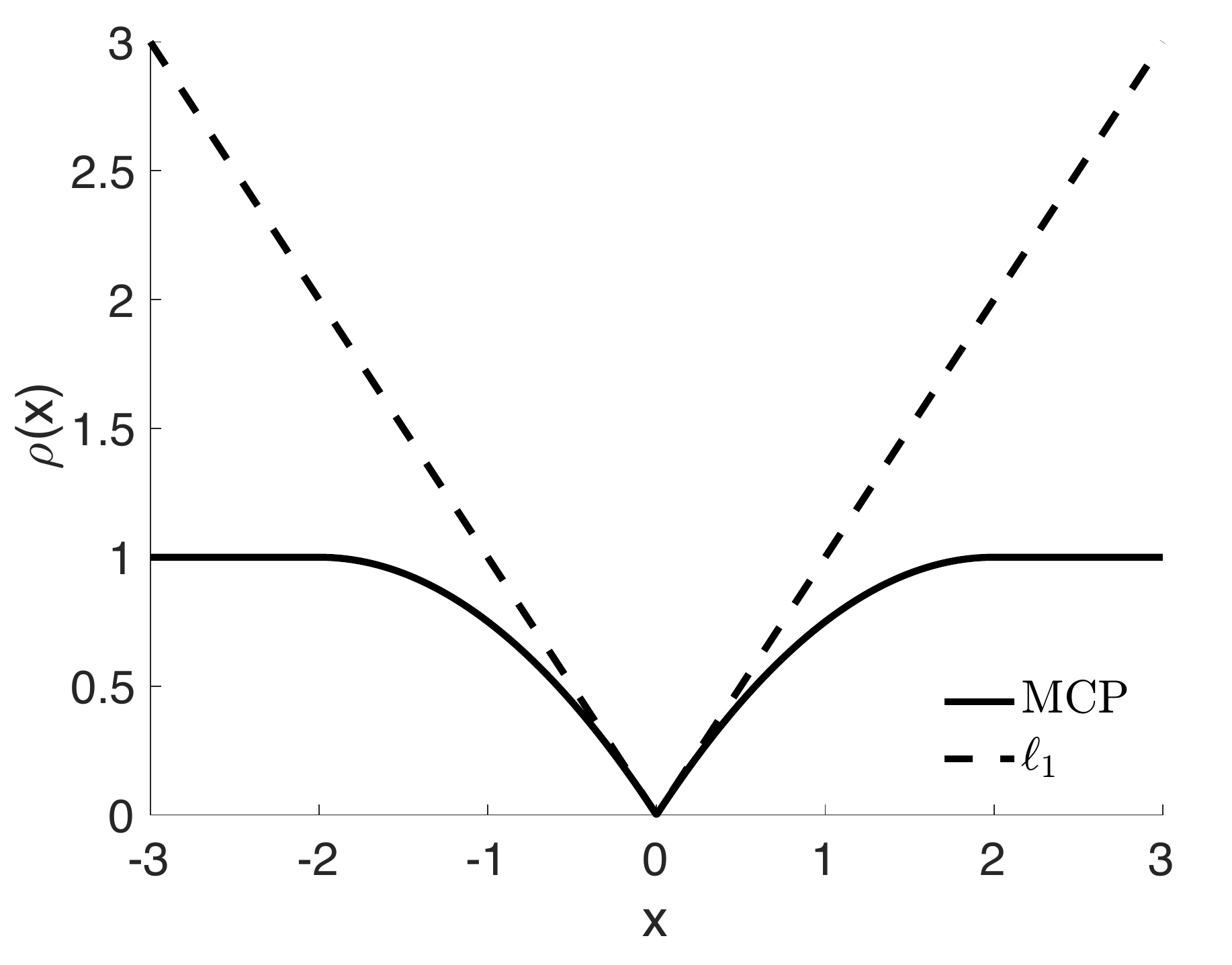}
	\caption{A comparison between the MCP (solid line) and the $\ell_1$ penalty (dashed line).} %The solid line is MCP with $(\gamma,\lambda) = (2,1)$. The dashed line is the $\ell_1$ penalty.}
	\label{fig:mcp}
\end{figure}

Throughout, let $\rho(\cdot) := \rho_\theta(\cdot)$ be the MCP with two parameters $\theta = (\gamma, \lambda)$~\citep{Z10_MCP}: 
\begin{align} \label{eq:mcp}
\rho (x; \gamma, \lambda) =
	\begin{cases} 
		\lambda |x| - \f {x^2} {2\gamma}, & \mbox{$|x| < \gamma\lambda$,}\\
		\f 1 2 \gamma\lambda^2, &\mbox{$|x|\geq \gamma\lambda$,}
	\end{cases}
\end{align}
where $\lambda \geq 0$ and $\gamma > 1$. Parameter $\lambda$ measures the penalty level, while $\gamma$ controls the concavity of the function. For a fixed value of $\lambda$, the MCP approaches the $\ell_1$ penalty as $\gamma \to \infty$, and the $\ell_0$ penalty as $\gamma \to 0^+$. 

Figure \ref{fig:mcp} compares the MCP with $(\gamma, \lambda) = (2,1)$ and the $\ell_1$ penalty. The right derivative of MCP at zero is $\lambda$, which is the same as the derivative of the $\ell_1$ penalty. The MCP function flats out when $|x| \geq \gamma  \lambda$. 

\begin{remark} \normalfont
Aragam and Zhou \cite{AZ15} use an MCP regularized likelihood to estimate Gaussian DAGs as well. However, rather than searching over permutations, which automatically satisfies the acyclicity constraint, they perform a greedy coordinate descent to minimize the regularized loss over DAGs. Thus, at each update in their algorithm the acyclicity constraint must be carefully checked.
\end{remark}

\subsection{Likelihood for experimental data} 

It is well-known that DAGs in the same Markov equivalence class are observationally equivalent, and thus we cannot distinguish such DAGs from observational data alone. However, experimental interventions can help distinguish equivalent DAGs and construct causal networks. Following~\cite{JP95}, intervention on a node $X_j$ in a DAG is to impose a fixed external distribution on this node, denoted by $p(x_j \mid \bullet)$, independent of all $X_{-j}$, while keeping the structural equations~\eqref{eq:sems} of the other nodes unchanged. 

Suppose that our data $\Xb\in \R^{n \times p}$ consists of $M$ blocks, where each block $\Xb^m \in \R^{n_m \times p}$ and $n = \summ m M n_m $. Denote by $X^m_\calI \subseteq \{X_1, \ldots, X_p\}$ the set of variables under experimental interventions in block $m$. Then, the data for $X_j \in X^m_\calI$ in this block are generated independently from the distribution $p(x_j\mid \bullet)$, while for $X_i \notin X_\calI^m$ from the conditional distribution $[X_i\mid\Pi_i^\calG]$. Note that multiple nodes could be intervened for a block of data, in which case $|X^m_\calI |\geq 2$.

Let $\calI_j \subseteq \{1,2,\ldots, n\}$ be the set of observations for which $X_j$  is under experimental intervention, and let $\calO_j  = \{1,2,\ldots, n\}\setminus {\calI_j}$ be its complement. 
By the truncated factorization formula~\citep{R86, JP95, SGS01_causal}, the joint density  of experimental data is
\begin{align} \label{eq:int:data}
	p(\Xb) = \prod_{j=1}^p \prod_{h\in\calO_j} p(x_{hj} \mid pa_{hj}) \prod_{k\in\calI_j}   p(x_{kj} \mid \bullet),  %\propto \prod_{j=1}^p \prod_{h\in\calO_j} p(x_{hj} \mid pa_{hj}),
\end{align}
where $x_{hj}$ is the value of  the $j^\text{th}$ variable in the $h^\text{th}$ observation and $pa_{hj}$ is the value for its parents. Let $\Xb_{\calO_{\pi(j)}}$ be the submatrix of $\Xb$ with rows in $\calO_{\pi(j)}$ and 
\begin{align*}
	\Sigh^{j} := \frac1{|\calO_{\pi(j)}|} \Xb_{\calO_{\pi(j)}}^\top \Xb_{\calO_{\pi(j)}}
\end{align*}
be  the sample covariance matrix computed from data in these rows. Then the log-likelihood of experimental data can be re-parametrized into the Cholesky loss functions as well:
\begin{lemma}\label{lem:interv:loglike}
	The negative log-likelihood for experimental data~\eqref{eq:int:data} can be written as
	\begin{align}
		\ell_{\calO}(L,P) 
		& = \summ j p |\calO_{\pi(j)}|  \lchol\left(L_j; P\Sigh^{j} P^\top \right),\label{eq:LP:ellO}
	\end{align} 
	where $L_j = (e_j - \beta_j) / {\omega_j} \in \R^p$, $L=(L_j)\in\calL_p$, and $|L_j| := L_{jj}$ in $\lchol(\cdot)$~\eqref{eq:chol:loss}.
\end{lemma}
See Appendix~\ref{proof:exp:like} for the proof. Though experimental data likelihood $\ell_{\calO}(L,P)$ in~\eqref{eq:LP:ellO} is not identical to the observational $\ell(L,P)$ in~\eqref{eq:LP:loss}, searching strategies described in the next section can be applied on both observational and experimental data.

%%%%%%%%%%%%%%%%%%%%%%%%%%%%%%%%%%%%%%%

\section{Optimization} \label{sec:optim}

We now describe how we solve the optimization problem~\eqref{eq:min_fp}. The main steps are outlined in Algorithm~\ref{algo:general}, where we use  simulated annealing to search over the permutation space for an ordering that minimizes the RC score defined in~\eqref{eq:fp}. To obtain the RC score for a given permutation, we need to solve a continuous optimization problem (line~\ref{alg:getl} and~\ref{alg:general:getl}) for which we propose a proximal gradient algorithm (Algorithm~\ref{algo:pg}).

\subsection{Searching over permutations}

\begin{algorithm}[t]
\caption{Annealing on regularized Cholesky score (ARCS).} \label{algo:general}
	\smallskip
\hspace*{\algorithmicindent} \textbf{Input:} Dataset $\Xb$, initial permutation matrix $P_0$, \par 
\hspace*{\algorithmicindent}{~~~~~~~~~~~} a temperature schedule $\{T^{(i)}, i = 0, \ldots, N\}$, \par
\hspace*{\algorithmicindent}{~~~~~~~~~~~} constant $m$.  \\
 \hspace*{\algorithmicindent} \textbf{Output:}  Adjacency matrix $\widehat B$.
 % \hspace*{\algorithmicindent} \textbf{Input:} $\Xb$, $P_0$, $m$, $\{T^{(i)}, i = 0, \ldots, N\}$. \\
 % \hspace*{\algorithmicindent} \textbf{Output:} $\widehat B$.
 \begin{algorithmic}[1]
	\State Select tuning parameters $(\gamma, \lambda)$ for $f(L;P)$ according to  % \par 
		 \noindent{} Algorithm~\ref{algo:bic} (Section \ref{sec:bic}). \label{algo:sarcs:bic}
	\State $\widehat P \gets P_0, ~ \widehat  L \gets \argmin_{L \in \calL_p} f(L;\widehat P)$ by Algorithm~\ref{algo:pg}, %\par 
		 \noindent{} $f(\widehat P) \gets f(\widehat L; \widehat P)$. \label{alg:getl}
	\For{$i = 0, \ldots, N$} 
		\State $T \gets T^{(i)}$.
		\State Propose $P^*$ by flipping a random length-$m$ interval \par 
			 \noindent{~~~~~} in the permutation defined by $\widehat P$.
		\State $L^* \gets \argmin_{L \in \calL_p} f(L;P^*)$ using Algorithm~\ref{algo:pg}, \par
			 \noindent{~~~~} $f(P^*) \gets f( L^*; P^*)$. \label{alg:general:getl}
		\State $\alpha \gets \min \left\{1, \exp\big({-}\frac1T[f(P^*)-f(\widehat P)]\big)\right\}$. \label{algo:alpha}
		\State Set $(\widehat P, \widehat L, f(\widehat P)) \gets (P^*, L^*, f(P^*))$ with prob. $\alpha$.
	\EndFor
	\State \textbf{end for} \label{algo1:end}
	\State Refine adjacency matrix $\widehat B$ given $(\widehat P, \widehat L )$ by Algorithm~\ref{algo:after-sa}  (Section \ref{sec:after-sa}). \label{algo:after-sa-line}
\end{algorithmic}
\end{algorithm}

\iffalse
\begin{figure*}[t]
\begin{center}
	\includegraphics[width=13cm] {figs/figure3.pdf}
	\caption{An operator on permutation that flips the interval (3,4,5,6).}
	\label{fig:per_oper_k}
\end{center}
\end{figure*}
\fi

The ARCS algorithm is detailed in Algorithm~\ref{algo:general}. At each iteration, we propose a permutation $P^*$ and decide whether to stay at the current permutation or move to the proposed one with probability $\alpha$ given in line~\ref{algo:alpha}. The probability is determined by the difference between the proposed and current scores $f(P^*) - f(\Ph)$ normalized by a temperature parameter $T$. For $T \to \infty$, the jumps are completely random and for $T \to 0^+$ completely determined by the RC score $f(\cdot)$. The algorithm follows a temperature schedule which is often taken to be a decreasing sequence $T^{(0)} \geq T^{(1)} \geq \ldots \geq T^{(N)}$ allowing the algorithm to explore more early on and zoom in on a solution as time progresses.

The proposed permutation matrix $P^*$ is constructed as follows. Let $\pih$ and $\pi^*$ be the permutations associated with $\Ph$ and $P^*$ as in~\eqref{eq:Ppi:def}. We propose $\pi^*$ by flipping (i.e., reversing the order of) a random interval of length $m$ in the current permutation $\pih$. For example, with $m=3$ we may flip $\pih =(1, 2, 3, 4, \ldots, p)$ to $\pi^* = (1, 4, 3, 2, \ldots, p)$ in the proposal. Equivalently, we flip a contiguous block of $m$ rows of $\Ph$ to generate $P^*$.

As a byproduct of evaluating the RC score for the proposed permutation $P^*$, we also obtain the corresponding lower triangular matrix $L^*$, representing the associated DAG. We keep track of these DAGs as well as the permutations throughout the algorithm (line~\ref{alg:general:getl}). 

\subsection{Computing RC score}

We propose a proximal gradient algorithm to evaluate the RC score $f(P)$ at each permutation matrix $P$ (line~\ref{alg:getl} and~\ref{alg:general:getl}, Algorithm~\ref{algo:general}). This algorithm belongs to a class of first-order methods that are quite effective at optimizing functions composed of a smooth loss and a nonsmooth penalty~\citep{proxAlgo13}.

The RC score is obtained by minimizing the RC loss $f(L;P)$ over $L$ as shown in~\eqref{eq:fp}. Recall that $\calL_p$ is the set of $p \times p$ lower triangular matrices, and let
\begin{align}\label{eq:tri:pen}
	\rho(u):=\sum_{i>j}\rho(u_{ij}), \quad \text{for} \quad u = (u_{ij}) \in \calL_p.
\end{align}
Note that we are leaving out the diagonal elements of $u$ in defining $\rho(u)$. 
% \aaa{This might be somewhat confusing. }
Then, the RC loss is $ f(u;P) = \ell(u,P) + \rho(u),$ where $\ell(u, P)$~\eqref{eq:LP:loss} is differentiable and $\rho(u)$ is nonsmooth. The idea of the proximal gradient algorithm is to replace $\ell(u,P)$ with a local quadratic function at the current estimate $L$ and optimize the resulting  approximation to $f(u;P)$ to get a new estimate $L^+$:
\begin{align} \label{eq:update_l}
	L^+&= \argmin_{u\in\calL_p} \ell(L) + \nabla \ell(L)^\top (u-L)  + \f {1}{2t} \lVert u-L \lVert^2 + \rho(u)  \nonumber\\
	& = \argmin_{u\in\calL_p}  \f 1 {2t} \norm{L-t\nabla \ell(L) - u}^2 + \rho(u) , 
\end{align}
where $\ell(L) = \ell(L,P), \nabla \ell (L) :=  \nabla_L \ell (L, P)$ is the gradient of $\ell (L, P)$ w.r.t. $L$, and $t>0$ is a step size. Consider the proximal operator $\prox_{\rho} : \calL_p \to \calL_p$ associated with $\rho$ defined by 
\[
\prox_{\rho}(x) := \argmin_{u \in \calL_p}\left(\rho(u) +\f 1 2 \lVert x-u\lVert^2\right),
\]
where  $x \in \calL_p$ and $\lVert \ \cdot \ \lVert$ is the usual Euclidean norm. Then, \eqref{eq:update_l} is equivalent to
\begin{align} \label{eq:l_update}
	L^+ = \prox_{ t \rho} \left(L - t \nabla \ell(L) \right),
\end{align}
where $\prox_{t\rho}(\cdot)$ is the proximal operator applied to the scaled function $t\rho(\cdot)$. Since $\rho(u) $ is separable across the coordinates $\{ u_{ij}, i\geq j\}$, we have for $x \in \calL_p$,
\begin{align*}
	\left(\prox_{\rho}(x)\right)_{ij} = 
	\begin{cases}
		\prox_{\rho}(x_{ij}), & i > j, \\
		\prox_{0}(x_{ii}) = x_{ii}, & i = j.
	\end{cases}
\end{align*}
The proximal operators on the RHS are univariate, and the distinction between the two cases is because we do not penalize the diagonal entries, i.e., $\rho(u_{ii}) = 0$.

\begin{algorithm}[t]
	\caption{Compute the RC score by proximal gradient.}
	\label{algo:pg}
	\hspace*{\algorithmicindent} \textbf{Input:} $P$, $L^{(0)} \in \calL_p$, $t^{(0)} > 0$, $\kappa \in (0,1)$, \texttt{max-iter},   
	\hspace*{\algorithmicindent}{~~~~~~~~~~~} \texttt{tol}. \\
	\hspace*{\algorithmicindent} \textbf{Output:}  $L$. 
	\begin{algorithmic}[1]
		\State $k \gets 0$, err $\gets \infty$, $L \gets L^{(0)}.$
		\While{$k < $ \texttt{max-iter} and {err} $>$ \texttt{tol}}
		\State Compute $\nabla \ell(L)$ using either Lemma~\ref{lemma:gradl} or~\ref{lemma:intv_grad}.
		\State $t \gets t^{(0)}/{\lVert \nabla \ell(L) \lVert_F} $. \label{algo:pg:step-size}
		% \Repeat
		\State \textbf{repeat}
		\State {~~~~}$\widetilde L \gets L -  t \nabla \ell(L)$.  \label{alg:l_start} 
		% \State {~} Update: $L^+_{ij} \gets \prox_{ t \rho} (\widetilde L_{ij})$ for $i>j$  (using Lemma~\ref{lemma:prox_mcp}), and $L^+_{ii} \gets \widetilde L_{ii}$. \label{algo:loff}
		 \State {~~~} $L^+_{ij} \gets \prox_{ t \rho} (\widetilde L_{ij})$ for $i>j$  (using Lemma~\ref{lemma:prox_mcp}).
		 \State {~~~} $L^+_{ii} \gets \widetilde L_{ii}$.
		\State {~~~} \textbf{break if} $\ell\left(L^+,P\right) \leq  \ell \left(L, P\right) + \left\langle {\nabla \ell (L)} , L^+ - L \right\rangle$ \par 
			\noindent{~~~~~~~~~~~~~~~~~~~~~~~~~~~~~~~~~~~~~~~~~~~~~~}  $+\frac1{2 t} \| L^+ - L\|^2_F$.\label{alg:pg_stopcon}
		\State {~~~} $t \gets \kappa t$.
		% \Until{$\ell\left(L^+,P\right) \leq \ell \left(L, P\right) + \left\langle {\nabla \ell (L)} , L^+ - L \right\rangle + \frac1{2 t} \| L^+ - L\|^2_F$.}\label{alg:pg_stopcon}
		\State {err}  
		$\gets \max_j \delta(L_j^+,L_j) $ where  $\delta(x,y) := \frac {\norm{x-y}} {\max\{1,\norm{y}\}}$.
		\State $L \gets L^+$ and $k \gets k+1$.
		\EndWhile
		\State \textbf{end while}
	\end{algorithmic}
\end{algorithm}

The overall procedure is summarized in Algorithm~\ref{algo:pg}. To choose the step size $t$ normalized by $\lVert \nabla \ell(L) \lVert_F$ (line~\ref{algo:pg:step-size}), we have used a line search strategy~\citep{proxAlgo13}, where we repeatedly reduce the step size by a factor $\kappa \in (0,1)$ until a quadratic upper bound is satisfied by the new update (line~\ref{alg:pg_stopcon}). To implement Algorithm \ref{algo:pg}, we need two more ingredients, $\nabla \ell(L)$ and the univariate $\prox_{\rho}(\cdot)$, both of which have nice closed-form expressions:

\begin{lemma} \label{lemma:gradl}
The gradient of  $\ell( L, P)$ in~\eqref{eq:LP:loss} w.r.t. $L$ is
\begin{align*}
	\nabla \ell (L) =n \left({\Pi_\calL}(P\widehat\Sigma P^\top L) - \diag\left(\{1/L_{ii}\}_{i=1}^p\right)\right),
\end{align*} 
where $\Pi_\calL: A \mapsto (A_{ij} \mathbbm{1}\{i \geq j\})_{p\times p}$ maps a matrix to its lower triangular projection.
\end{lemma} 

\begin{lemma} \label{lemma:intv_grad}
The gradient of $\ell_\calO(L,P)$ in~\eqref{eq:LP:ellO} w.r.t. $L_j$ is
\[ 
\nabla_{L_j} \ell_{\calO}(L,P) = \left|\calO_{\pi(j)}\right|  \left(\Pi_j\big(P\Sigh^{j}P^\top L_j\big) - \frac {e_j} {L_{jj}}\right),
\] 
where $\Pi_j : v \mapsto (v_{i} \mathbbm{1}\{i \geq j\})_{p\times 1}$ and $\{e_j\}$ is the canonical basis of $\R^p$.
\end{lemma}

\begin{lemma} \label{lemma:prox_mcp}
Let $\rho$ be the scalar MCP with parameter $(\gamma, \lambda)$ defined in~\eqref{eq:mcp}, and let $\rho_1$ be the same penalty for $\lambda = \gamma = 1$. Then, for any $t > 0$,
\begin{equation} \label{eq:prox-lemma}
	\prox_{t \rho}( x) =\lambda \gamma  \prox_{(t/\gamma) \rho_1}\Big( \frac{x}{\lambda \gamma}\Big), 
\end{equation} 
and for any $\alpha  > 0$, 
\begin{align} \label{eq:prox-mcp-formula}
	\prox_{\alpha \rho_1}(x) = 
		\begin{cases}
			0, & \parbox[c]{.25\textwidth}{ $0 \leq x < \min\{\alpha,1\} \text{ or } \\ 1 < x < \sqrt\alpha;$}\\
			\dfrac{x-\alpha}{1-\alpha}, & \alpha < x \leq 1;\\
			x, & \parbox[c]{.25\textwidth}{$x > \max\{\alpha,1\} \text{ or } \\ 1 < \sqrt\alpha < x \leq \alpha.$}
		\end{cases}
\end{align}
Moreover, $\prox_{\alpha \rho_1}(-x) = - \prox_{\alpha \rho_1}(x)$ for all $x \in \mathbb R$.
\end{lemma}

We have excluded two special cases in \eqref{eq:prox-mcp-formula} in which the minimizer is not unique: 
1) If $x = \alpha = 1$, $\prox_{\alpha \rho_1}(x) = [0,1]$; 
2) If $x =\sqrt\alpha >1$, $\prox_{\alpha \rho_1}(x) = \{0, \sqrt\alpha\}$. We set $\prox_{\alpha \rho_1}(x) = 0$ in our implementation if these special cases occur. The MCP has parameter $\gamma > 1$, and usually the step size $t < 1$. Thus, the cases with $\alpha < 1$ are the most common scenario in our numerical study. These lemmas are proved in Appendix~\ref{proof:lemma:gradl},~\ref{proof:intv_grad}, and~\ref{app:prox_mcp}.

\subsection{Structure refinement after annealing} \label{sec:after-sa}

At the end of the annealing loop (line~\ref{algo1:end}, Algorithm~\ref{algo:general}), a pair $(\widehat P, \widehat L)$ is found which minimizes the RC score~\eqref{eq:min_fp}. 
Accordingly, an estimated reversal of a topological sort is $\widehat \pi=\widehat P(1,\ldots,p)^\top$. Define $\widetilde L = \Ph^\top \widehat L \Ph$, and $\widehat B$ by $\widehat B_{ij} = - {\widetilde L_{ij}} / {\widetilde L_{jj}}$ for $i\neq j$ and $\widehat B_{ii} = 0$. Then, $\widehat B$ is the estimated weighted adjacency matrix for a DAG, i.e., an estimate for $B_0$. The support of $\widehat B$ gives the estimated parent sets $\widehat{pa}_j= \{ i: \widehat B_{ij}\neq 0\}$ for $j=1,\ldots,p$. The use of a continuous regularizer, i.e. MCP, eases our optimization problem; however, this may lead to more false positive edges compared to $\ell_0$ regularization. To improve structure learning accuracy, we add a refinement step to remove some predicted edges by conditional independence tests, which borrows the strength from a constraint-based approach.

The refinement step outlined in Algorithm~\ref{algo:after-sa} is based on the following fact. If $k\prec j$ in a topological sort and there is no edge $k\to j$, then $X_k \perp X_j \mid \Pi_j$, where $\Pi_j$ is the parent set of $X_j$. For each $k \in \widehat{pa}_j$, we test the null hypothesis that $X_k$ and $X_j$ are conditionally independent given $\widehat {pa}_j \setminus \{ k \}$ using the Fisher Z-score. We remove the edge $k\to j$ if the null hypothesis is not rejected at a given significance level. The conditional independence tests are performed in a sequential manner for the nodes in $\widehat{pa}_j$ according to the estimated topological sort: For $k_1,k_2 \in \widehat{pa}_j$, if $k_1\prec k_2$ in the sort, we carry out the test for $k_2$ prior to that for $k_1$.

\begin{algorithm}[t]
\caption{Constraint-based structure refinement.} \label{algo:after-sa}
\hspace*{\algorithmicindent} \textbf{Input:} Dataset $\Xb$, permutation ${\widehat \pi}$, adjacency matrix $\widehat B$, \\
\hspace*{\algorithmicindent}{~~~~~~~~~~~~}significance level $\alpha$. \\  
\hspace*{\algorithmicindent} \textbf{Output:}  Adjacency matrix $\widehat B$.
\begin{algorithmic}[1]	
	\State $Z_\alpha \gets \Phi^{-1}(1-\f \alpha 2)$, where $\Phi(x)$ is the CDF of $\calN(0,1)$.
	\For{$j = 1, \ldots, p$} 
		\State $\widehat {pa}_j \gets \{ i: \widehat B_{ij}\neq 0\}$. 
		\For{$ k \in \widehat {pa}_j $}
			\State $\mathbf{s} \gets \widehat {pa}_j  \setminus \{k\}$.
			\State $\Xb^j \gets $ observations for which $j$ is not intervened
			\State $n \gets$ number of rows in $\Xb^j$.
			\State $r_{j,k \mid \mathbf{s}} \gets$ sample partial correlation between $X_j$ \par 
			\noindent{~~~~~~~~~} and $X_k$ given $X_\mathbf{s}$ based on $\Xb^j$.
			\State $z\gets \f 1 2 \sqrt{n-|\mathbf{s}|-3} \log \left(\f {1+r_{j,k \mid \mathbf{s}}} {1-r_{j,k \mid \mathbf{s}}}\right)$. 
			\State Remove $k$ from $\widehat {pa}_j$, if $|z| < Z_\alpha$. 
		\EndFor
		\State \textbf{end for}
		\State $\widehat B_{ij} \gets 1$ if $i \in \widehat {pa}_j$ and $\widehat B_{ij} \gets 0$ otherwise.
	\EndFor
\State \textbf{end for}
\end{algorithmic}
\end{algorithm}

\subsection{Selection of the tuning parameters}\label{sec:bic}

Before starting the iterations in Algorithm~\ref{algo:general}, we select and fix the tuning parameters $\theta = (\gamma, \lambda)$ of MCP (line~\ref{algo:sarcs:bic}), hence fixing a particular scoring function $f(L,P) = f_\theta(L,P)$ throughout the algorithm. 

\begin{algorithm}[t]
\caption{Tuning parameter selection by BIC.}\label{algo:bic}
\hspace*{\algorithmicindent} \textbf{Input:} Initial permutation $P_0$ and a grid of values \\
\hspace*{\algorithmicindent}{~~~~~~~~~~~}  $\{\theta^{(i)}\} = \{(\gamma^{(i)},\lambda^{(i)})\}$. \\
\hspace*{\algorithmicindent} \textbf{Output:}  Optimal index $i^*$ in the grid.
\begin{algorithmic}[1]
	\State Define $\widehat L(\theta) := \argmin_{L\in\calL_p} \,f_{\theta}(L;P_0)$ computed by  Algorithm~\ref{algo:pg}.
	\State Let $\text{BIC}(\theta) := 2 \ell\big( \widehat L(\theta) ; P_0\big) + \|
			\widehat L(\theta) \|_0 \log\left(\max\{n,p\}\right).$ \label{algo:bic:val}
	\State Output $i^* = \argmin_{i} \text{BIC}(\theta^{(i)})$.
\end{algorithmic}
\end{algorithm}

We use the Bayesian information criterion (BIC)~\citep{S78} to select the tuning parameters, given an initial permutation $P_0$.  The details are summarized in Algorithm~\ref{algo:bic}. For every pair $(\gamma^{(i)},\lambda^{(i)})$ over a grid of values, we evaluate the BIC score given in line~\ref{algo:bic:val}, where $\|\widehat L(\cdot) \|_0$ is the number of nonzero entries in $\widehat L(\cdot)$, and then we output the one with the lowest BIC score. The regularization parameter in BIC($\theta$) is adapted to $\log(\max\{n,p\})$, which works well for both low and high-dimensional data.
To construct the grid, possible choices for the concavity parameter $\gamma$ are $\{2, 10, 50, 100\}$ based on our tests. Note that $\gamma>1$ is required in the definition of MCP \eqref{eq:mcp}, while the behavior of MCP for $\gamma\geq 100$ is essentially the same as the $\ell_1$ penalty. For the regularization parameter $\lambda$, we select 20 equi-spaced points from the interval $[0.1 \sqrt{n},\sqrt{n}]$. The choice of $\sqrt{n}$  often leads to an empty graph when the data are standardized, hence a natural end point.

%%%%%%%%%%%%%%%%%%%%%%%%%%%%%%%%%%%%%%%
\section{Results}\label{sec:res}

\subsection{Methods and data}

Recall that $p$ is the number of variables and $n$ is the number of observations. For a thorough evaluation of the algorithm, we simulated data for both $n>p$ and $n<p$ cases.

We used real and synthetic networks to simulate data. Real networks were downloaded from the Bayesian networks online repository~\citep{bnrep}. We duplicated some of them to further increase the network size. Synthetic DAG structures were constructed using the \texttt{sparsebn} package \citep{AGZ}. Given a DAG structure, we sampled the edge coefficients $\beta_{ij}$ uniformly from $[-0.8, -0.5] \cup [0.5,0.8]$ and set the noise variance to one. We then calculated the covariance matrix according to~\eqref{eq:sigma_0} and normalized its diagonal elements to one. Consequently, the variances of $\{X_1, \ldots, X_p\}$ were identical. We used the following networks to generate observational data, denoted by the network name and $(p,s_0)$, where $s_0$ is the number of edges after duplication: 4 copies of \texttt{Hailfinder} (224, 264), 1 copy of \texttt{Andes} (223,  338), 2 copies of \texttt{Hepar2}  (280,  492), 4 copies of \texttt{Win95pts} (304, 448), 1 copy of \texttt{Pigs} (441, 592), and random DAGs, \texttt{rDAG1} (300, 300) and \texttt{rDAG2} (300, 600).

In the observational data setting, we compared our algorithm with the following BN learning algorithms: the coordinate descent (CD) algorithm \citep{AZ15}, the standard greedy hill climbing (HC) algorithm \citep{GMP11}, the greedy equivalence search (GES) \citep{C02}, the Peter-Clark (PC) algorithm \citep{pc_91}, the max-min hill-climbing (MMHC) algorithm \citep{TEA06}, and the genetic algorithm (GA) \citep{CPV18}. 

The CD algorithm optimizes a regularized log-likelihood function by a blockwise update on $(\beta_{ij}, \beta_{ji})$ while checking the acyclicity constraint before each update.
The HC algorithm performs a greedy search over the DAG space by starting from a certain initial state, performing a finite number of local changes and selecting the DAG with the best improvement in each local change. 
The GES algorithm searches over the equivalence classes and utilizes greedy search operators on the current state to find the next one, of which the output is an equivalence class of DAGs. 
The PC algorithm performs conditional independence tests to identify edges and orients edge directions afterwards. 
The MMHC algorithm constructs the skeleton of a Bayesian network via conditional independence tests and then performs a greedy hill climbing search to orient the edges via optimizing a Bayesian score. 
The GA decomposes graph estimation into two optimization sub-problems: node ordering search with mutation and crossover operators, and structure optimization by an adaption of the least angle regression~\citep{EHJT04}.

Among these methods, PC is a constraint-based method, and MMHC is a hybrid method. Other methods, CD, HC, GES and GA, are all score-based, where CD and HC search over the DAG space, GES searches over the equivalence classes, and GA searches over the permutation space. Our method is a score-based search over the permutation space, similar to GA. 

Our ARCS algorithm (Algorithm~\ref{algo:general}) may take an initial permutation $P_0$ provided by a local search method. In this study, we use the CD and GES algorithms to provide an initial permutation, and call the corresponding implementation ARCS(CD) and ARCS(GES). To partially preserve properties of the input initial permutation, we start with a low temperature $T^{(0)}=1$. The output of the CD algorithm is a DAG for which we find a topological sort to define $P_0$. The GES algorithm outputs a completed partially directed acyclic graph (CPDAG). We then generate a DAG in the equivalence class of the estimated CPDAG, and initialize ARCS with a topological sort of this DAG.

We implemented the ARCS algorithm in \MATLAB, and used the following R packages for other methods: \texttt{sparsebn}~\citep{AGZ}  for the CD algorithm,  \texttt{rcausal}~\citep{J15} for the GES, GIES (for experimental data) and PC algorithms, \texttt{bnlearn}~\citep{S10} for the MMHC and HC algorithms, and \texttt{GADAG}~\citep{CPV18} for the GA. 

\subsection{Accuracy metrics}

Among all methods applied on observational data, ARCS, CD, HC and MMHC output DAGs, while the GES and PC algorithms output CPDAGs. Given these estimates, we need to evaluate the performance of each method.  Define P, TP, FP, M, R as the numbers of estimated edges, true positive edges, false positive edges, missing edges and reverse edges, respectively. To standardize the performance metrics in observational data setting, we consider both directed and undirected edges in the definitions of these metrics as follows.

P is the number of edges in the estimated graph. FP is the number of edges in the estimated graph skeleton but not in the true skeleton. M counts the number of edges in the true skeleton but not in the skeleton of the estimated graph. We define an estimated directed edge to be true positive (TP) if it meets either of the two criteria: 
1) This edge is in the true DAG with the same orientation; 
2) This edge coincides after converting the estimated graph and true DAG to CPDAGs. 
An estimated undirected edge is considered TP if it satisfies the second condition. Note that our criterion takes into account reversible edges in an equivalence class for observational data. Lastly, the number of reversed edges $\text{R} = \text{P} - \text{TP} - \text{FP}$. 

Denote by $s_0$ the number of edges in the true DAG. The overall accuracy of a method is measured by the structural Hamming distance (SHD) and Jaccard index (JI), where $\text{SHD} = \text{R} + \text{FP} + \text{M}$ and $\text{JI} = \text{TP} / (s_0 + \text{P} - \text{TP})$. A method has better performance if it achieves a lower SHD and/or a higher JI.

\subsection{Comparison on observational data}

We used large networks, where $p \in (200, 450)$ and $s_0 \in (250, 600]$, to simulate observational data with $n<p$ and $n>p$. For each setting $(p, s_0, n)$, we generated 20 datasets, and ran CD, ARCS(CD), GES, ARCS(GES) and other methods (PC, HC, MMHC and GA) with a maximum time allowance of 10 minutes per dataset. The HC and MMHC algorithms had an upper-bound of the in-degree number as 2. We tried a higher maximum in-degree, but it resulted in a large FP. MMHC and PC were run with a significance level of 0.01 in conditional independence tests. We ran the CD algorithm with an MCP regularized likelihood, in which $\gamma = 2$ and $\lambda$ was chosen by a default model selection mechanism. GA was run for a maximum of $10^4$ iterations, using the default population size and the default rates of mutation and crossover. We tried a larger population size for GA, but it was too time-consuming.

Our methods, ARCS(CD) and ARCS(GES), initialized with permutations from CD and GES estimates, were run for a maximum of $10^4$ iterations, with initial temperature $T^{(0)}=1$  and reversal length $m=4$ (Algorithm~\ref{algo:general}). A $p$-value cutoff of $10^{-5}$ was used in the refinement step (Algorithm~\ref{algo:after-sa}). For the networks we considered, on average, 500 tests were performed in the refinement step, and the cutoff was chosen by Bonferroni correction to control the familywise error rate at level 0.005. In fact, our results were almost identical for any $p$-value cutoff between $10^{-3}$ and $10^{-5}$.

\begin{table}[t!]
\setlength{\tabcolsep}{4pt} 
\centering
\caption{Comparison between ARCS and initial estimates on observational data ($n<p$).} \label{tab:n<p:imp}
\medskip
\scalebox{0.9}{
\begin{threeparttable}
\begin{tabular}{ccrrrrrrr}
 	\hline
    	network & method & P & TP & R &FP & SHD & JI \\ 
	$(p, s_0, n)$ &&&&&\\
 	\hline
    	\texttt{Hailfinder}    
   	& CD & 270 & 154 & 84 & 32 & 142 & 0.41 \\ 
	$(224, 264, 200)$  & ARCS(CD) & 274 & 187 & 54 & 33 & 110 & 0.53 \\ 
   	& GES & 243 & 184 & 49 & 10 & 89 & 0.57 \\
   	% & ARCS(GES) & 279.2 & 210.9 & 37.4 & 30.9 & {\textbf{83.9}} & {\textbf{0.64}}
	& ARCS(GES) & 261 & 215 & 30 & 16 & {\textbf{65}} & {\textbf{0.70}}\vspace{1mm}\\
   	\texttt{Andes} & CD & 352 & 176 & 108 & 68 & 230 & 0.34 \\ 
   	$(223, 338, 200)$ & ARCS(CD) & 368 & 233 & 71 & 65 & 170 & 0.50 \\ 
   	& GES & 272 & 220 & 35 & 17 & 135 & 0.57 \\
   	% & ARCS(GES) & 334.7 & 276.7 & 30.7 & 27.3 & {\textbf{88.6}} & {\textbf{0.70}}
	& ARCS(GES) & 334 & 276 & 31 & 27 & {\textbf{89}} & {\textbf{0.70}}\vspace{1mm}\\
   	\texttt{Hepar2}
   	& CD & 495 & 245 & 120 & 130 & 377 & 0.33 \\ 
   	(280,492,200)  & ARCS(CD) & 500 & 311 & 107 & 82 & 264 & 0.46 \\ 
   	& GES & 410 & 253 & 96 & 62 & 302 & 0.39 \\ 
   	% & ARCS(GES) & 487.1 & 322.9 & 96.8 & 67.3 & {\textbf{236.3}} & {\textbf{0.49}}\vspace{1mm}\\
	& ARCS(GES) & 483 & 324 & 96 & 64 & {\textbf{232}} & {\textbf{0.50}}\vspace{1mm}\\
   	\texttt{Win95pts}   
   	& CD & 399 & 195 & 154 & 50 & 302 & 0.30 \\ 
    	$(304, 448, 200)$ & ARCS(CD) & 465 & 317 & 88 & 59 & 190 & 0.53 \\ 
   	& GES & 335 & 244 & 71 & 21 & 225 & 0.45 \\ 
   	% & ARCS(GES) & 449.8 & 362.0 & 56.4 & 31.4 & {\textbf{117.4}} & {\textbf{0.68}}
	& ARCS(GES) & 451 & 361 & 57 & 34 & {\textbf{121}} & {\textbf{0.67}}\vspace{1mm}\\
   	$\texttt{Pigs}$\footnotemark
   	& CD & 715 & 342 & 207 & 166 & 417 & 0.36 \\ 
    	$(441, 592, 200)$ & ARCS(CD) & 587 & 421 & 123 & 43 & 214 & 0.55 \\ 
    	& GES & 581 & 431 & 112 & 38 & 199 & 0.58 \\ 
    	% & ARCS(GES) & 565.5 & 451.4 & 91.7 & 22.5 & {\textbf{163.1}} & {\textbf{0.64}}
	& ARCS(GES) & 575 & 454 & 94 & 27 & {\textbf{165}} & {\textbf{0.63}}\vspace{1mm}\\
    	\texttt{rDAG1} 
    	& CD & 319 & 208 & 85 & 25 & 117 & 0.51 \\
    	$(300, 300, 240)$  
    	& ARCS(CD) & 305 & 260 & 36 & 9 & 49 & 0.76 \\ 
    	& GES & 283 & 268 & 14 & 1 & 33 & 0.85 \\ 
    	% & ARCS(GES) & 299.5 & 281.6 & 15.1 & 2.8 & {\textbf{21.1}} & {\textbf{0.89}}\vspace{1mm} \\
	& ARCS(GES) & 297 & 283 & 13 & 1 & {\textbf{18}} & {\textbf{0.90}}\vspace{1mm} \\
    	\texttt{rDAG2} 
    	& CD & 715 & 407 & 136 & 171 & 364 & 0.50 \\ 
    	$(300, 600, 240)$ 
    	& ARCS(CD) & 623 & 498 & 73 & 52 & 154 & 0.69 \\ 
    	& GES & 505 & 474 & 26 & 6 & 132 & 0.75 \\ 
    	% & ARCS(GES) & 595.9 & 561.5 & 19.8 & 14.7 & {\textbf{53.2}} & {\textbf{0.89}}\\ 
	& ARCS(GES) & 594 & 564 & 18 & 13 & {\textbf{49}} & {\textbf{0.89}} \\ 
  	\hline
	\end{tabular}
	\begin{tablenotes}[flushleft, para]
	In this and all subsequent tables, reported results are rounded averages over 20 datasets; the best SHD and JI for each network are highlighted in boldface.
	\end{tablenotes}
	\end{threeparttable}}
\end{table}

\footnotetext{Results for \texttt{Pigs} are averaged over 17 datasets because CD estimates were too dense using the default model selection criterion for the other three datasets.}

% ARCS vs CD, GES
\noindent{\bf ARCS versus CD and GES.} Table \ref{tab:n<p:imp} reports the average performance metrics across 20 datasets for 7 networks (5 real and 2 random networks) in the high-dimensional setting $n<p$ using  CD, ARCS(CD), GES and ARCS(GES). We were interested in the potential improvement of ARCS upon its initial permutations. It is indeed confirmed by the results in the table that ARCS(CD) and ARCS(GES) outperformed CD and GES, respectively, for every network, achieving lower SHDs and higher JIs. The reduction in SHD was very substantial, close to or above 40\%, for many networks, such as \texttt{Win95pts}, \texttt{Pigs}, \texttt{rDAG1} and \texttt{rDAG2}. 

The same comparison was done for low-dimensional settings with $n>p$, reported in Table~\ref{tab:n>p:imp}. We observe similar improvements of ARCS(CD) and ARCS(GES) over CD and GES, consistent with the results for high-dimensional data. It is seen from both tables that ARCS always increased TP, while maintaining or slightly reducing FP. The annealing process often identified more TP edges, while the refinement step (Algorithm~\ref{algo:after-sa}) cut down the FP edges given the ordering and parent sets learned through simulated annealing.

\begin{table}[t]
\setlength{\tabcolsep}{4pt} 
\centering
\caption{Comparison between ARCS and initial estimates on observational data ($n>p$).} \label{tab:n>p:imp}
\scalebox{0.9}{
\begin{tabular}{ccrrrrrr}
\hline
	network & method & P & TP & R & FP & SHD & JI \\ 
	$(p, s_0, n)$ &&&&&\\
	\hline
	\texttt{Hailfinder} 
   	& CD & 278 & 156 & 88 & 34 & 142 & 0.41 \\ 
 	$(224, 264, 400)$   & ARCS(CD) & 296 & 196 & 59 & 41 & 109 & 0.54 \\ 
   	& GES & 272 & 201 & 53 & 18 & 81 & 0.60 \\ 
   	% & ARCS(GES) & 300.1 & 212.8 & 40.9 & 46.5 & 97.7 & {\textbf{0.61}}\vspace{1mm}\\
	& ARCS(GES) & 273 & 232 & 26 & 15 & {\textbf{47}} & {\textbf{0.76}}\vspace{1mm}\\
	\texttt{Andes} 
   	& CD & 366 & 189 & 107 & 70 & 220 & 0.37 \\ 
 	$(223, 338, 400)$   & ARCS(CD) & 376 & 242 & 69 & 66 & 162 & 0.53 \\
   	& GES & 342 & 273 & 33 & 36 & 102 & 0.67 \\ 
   	% & ARCS(GES) & 342.9 & 293.9 & 24.8 & 24.1 & {\textbf{68.2}} & {\textbf{0.76}}\vspace{1mm} \\
	& ARCS(GES) & 348 & 296 & 26 & 26 & {\textbf{69}} & {\textbf{0.76}}\vspace{1mm} \\
  	\texttt{Hepar2}  
     	& CD & 519 & 267 & 130 & 122 & 347 & 0.36 \\ 
  	(280,492,400)  & ARCS(CD) & 523 & 330 & 106 & 87 & 248 & 0.49 \\ 
   	& GES & 509 & 312 & 113 & 84 & 264 & 0.45 \\ 
   	% & ARCS(GES) & 508.1 & 339.8 & 96.8 & 71.5 & {\textbf{223.8}} & {\textbf{0.52}}\vspace{1mm} \\
	& ARCS(GES) & 505 & 336 & 95 & 75 & {\textbf{231}} & {\textbf{0.51}}\vspace{1mm} \\
   	\texttt{Win95pts} 
   	& CD & 398 & 205 & 154 & 39 & 282 & 0.32 \\ 
  	$(304, 448, 500)$  & ARCS(CD) & 542 & 334 & 97 & 111 & 225 & 0.51 \\ 
   	& GES & 472 & 332 & 76 & 64 & 180 & 0.56 \\ 
   	% & ARCS(GES) & 473.3 & 391.6 & 45.7 & 36.0 & {\textbf{92.3}} & {\textbf{0.74}}
	& ARCS(GES) & 475 & 390 & 47 & 38 & {\textbf{96}} & {\textbf{0.73}}\vspace{1mm} \\
  	\texttt{Pigs}  
   	& CD & 755 & 353 & 224 & 179 & 418 & 0.36 \\ 
  	$(441, 592, 600)$ & ARCS(CD) & 631 & 451 & 123 & 57 & 198 & 0.59 \\ 
   	& GES & 645 & 467 & 122 & 57 & 182 & 0.61 \\
   	% & ARCS(GES)  & 612.4 & 476.0 & 97.7 & 38.7 & {\textbf{154.7}} & {\textbf{0.66}}\vspace{1mm} \\
	& ARCS(GES) & 619 & 471 & 102 & 46 & {\textbf{166}} & {\textbf{0.64}}\vspace{1mm} \\
 	\texttt{rDAG1} 
	& CD & 319 & 211 & 84 & 24 & 113 & 0.52 \\ 
  	$(300, 300, 450)$     & ARCS(CD) & 319 & 264 & 36 & 20 & 56 & 0.74 \\ 
   	& GES & 299 & 283 & 14 & 2 & 19 & 0.90 \\ 
   	% & ARCS(GES) & 301.5 & 283.5 & 15.7 & 2.4 & {\textbf{18.9}} & 0.89\vspace{1mm} \\
	& ARCS(GES) & 300 & 287 & 12 & 1 & {\textbf{14}} & {\textbf{0.92}}\vspace{1mm} \\
 	\texttt{rDAG2} 
   	& CD & 605 & 403 & 140 & 62 & 258 & 0.50 \\ 
  	$(300, 600, 600)$ 
  	& ARCS(CD) & 696 & 510 & 79 & 106 & 196 & 0.65 \\ 
   	& GES & 596 & 559 & 20 & 17 & 57 & 0.88 \\ 
    	% & ARCS(GES) & 608.3 & 580.6 & 16.2 & 11.4 & {\textbf{30.8}} & {\textbf{0.93}} \\
	& ARCS(GES) & 608 & 583 & 14 & 11 & {\textbf{28}} & {\textbf{0.93}} \\ 
  	\hline
\end{tabular}}
\smallskip
\end{table}

\begin{figure}%[t] 
\centering
	\includegraphics[width=.49\textwidth]{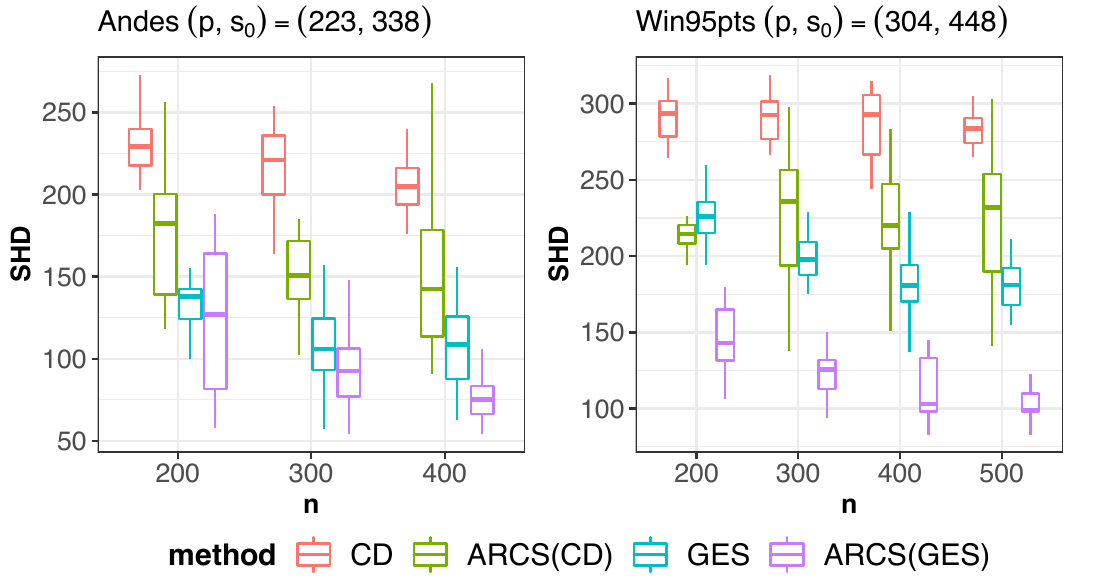}
	\caption{Boxplots of SHDs for CD, ARCS(CD), GES and ARCS(GES) on observational data from two networks.} % {\em Left:} Network \texttt{Andes} with $n \in  \{200, 300, 400\}$. {\em Right:} Network \texttt{Win95pts} with $n \in \{200, 300, 400, 500\}$.}
	\label{fig:obs_exmp} 
\end{figure}

Figure \ref{fig:obs_exmp} shows the clear improvement of the ARCS algorithm over CD and GES for the \texttt{Andes} and \texttt{Win95pts} networks with more choices of the sample size $n$. Each boxplot summarizes the SHDs over 20 datasets (excluding outliers). For all $n$, ARCS(GES) had a lower SHD than GES, and ARCS(CD) had a lower SHD than CD. The SHD distributions were well-separated, supporting the conclusion that ARCS significantly improves the accuracy of its initial estimates.

\begin{table}[t!]
\setlength{\tabcolsep}{4pt} 
	\centering
	\caption{ARCS against others on observational data ($n<p$).}
	\label{tab:n<p}
	\medskip
	\scalebox{0.9}{
	\begin{threeparttable}
		\begin{tabular}{ccrrrrrrr}
			\hline
			{network}  & method & P & TP & R & FP & SHD & JI \\ 
			$(p, s_0, n)$ & & & & & \\
			\hline
			\texttt{Hailfinder} 
			% & ARCS(GES) & 279.2 & 210.9 & 37.4 & 30.9 & {\textbf{83.9}} & {\textbf{0.64}}\\
			& ARCS(GES) & 261 & 215 & 30 & 16 & {\textbf{65}} & {\textbf{0.70}} \\
			$ (224, 264, 200)$    & HC & 414 & 120 & 116 & 179 & 324 & 0.21 \\ 
			& PC & 227 & 127 & 83 & 17 & 154 & 0.35 \\ 
			& GA & 193 & 68 & 54 & 71 & 266 & 0.18\vspace{1mm} \\
			\texttt{Andes}   
			% & ARCS(GES) & 334.7 & 276.7 & 30.7 & 27.3 & {\textbf{88.6}} & {\textbf{0.70}}\\
			& ARCS(GES) & 334 & 276 & 31 & 27 & {\textbf{89}} & {\textbf{0.70}}\\
			$ (223, 338, 200)$  & HC & 422 & 140 & 125 & 158 & 356 & 0.23 \\ 
			& MMHC & 257 & 146 & 98 & 13 & 205 & 0.33 \\ 
			& PC & 276 & 120 & 137 & 18 & 236 & 0.24 \\
			& GA & 214 & 77 & 67 & 70 & 331 & 0.16\vspace{1mm} \\
			\texttt{Hepar2}
			% & ARCS(GES) & 487.1 & 322.9 & 96.8 & 67.3 & {\textbf{236.3}} & {\textbf{0.49}}\\
			& ARCS(GES) & 483 & 324 & 96 & 64 & {\textbf{232}} & {\textbf{0.50}} \\ 
			$ (280,492,200)$ 
			& HC & 540 & 153 & 175 & 212 & 551 & 0.17 \\ 
			& PC & 310 & 108 & 149 & 53 & 436 & 0.16 \\ 
			& MMHC & 269 & 108 & 135 & 26 & 410 & 0.17 \\ 
			& GA   & 434 & 103 & 85 & 246 & 636 & 0.10\vspace{1mm} \\
			\texttt{Win95pts}  
			% & ARCS(GES) & 449.8 & 362.0 & 56.4 & 31.4 & {\textbf{117.4}} & {\textbf{0.68}} \\ 
			& ARCS(GES) & 451 & 361 & 57 & 34 & {\textbf{121}} & {\textbf{0.67}} \\ 
			$ (304, 448, 200)$ & HC & 581 & 140 & 206 & 235 & 544 & 0.16 \\ 
			& MMHC & 342 & 146 & 172 & 25 & 327 & 0.23 \\ 
			& PC & 369 & 114 & 221 & 35 & 369 & 0.16 \\ 
			& GA & 255 & 108 & 87 & 60 & 400 & 0.18\vspace{1mm} \\
			\texttt{Pigs} 
			% & ARCS(GES) & 565.5 & 451.4 & 91.7 & 22.5 & {\textbf{163.1}} & {\textbf{0.64}} \\ 
			& ARCS(GES) & 575 & 454 & 94 & 27 & {\textbf{165}} & {\textbf{0.63}} \\ 
			$ (441, 592, 200)$ & HC & 856 & 367 & 168 & 322 & 546 & 0.34 \\ 
			& GA & 437 & 135 & 127 & 175 & 632 & 0.15\vspace{1mm} \\
			\texttt{rDAG1} 
			% & ARCS(GES) & 299.5 & 281.6 & 15.1 & 2.8 & {\textbf{21.1}} & {\textbf{0.89}} \\ 
			& ARCS(GES) & 297 & 283 & 13 & 1 & {\textbf{18}} & {\textbf{0.90}} \\ 
			$ (300, 300, 240)$  & HC & 562 & 135 & 155 & 273 & 438 & 0.19 \\ 
			& MMHC & 316 & 143 & 144 & 28 & 188 & 0.30 \\ 
			& PC & 328 & 166 & 126 & 36 & 170 & 0.36 \\
			& GA & 175 & 89 & 56 & 30 & 242 & 0.23\vspace{1mm} \\
			\texttt{rDAG2}  
			% & ARCS(GES) & 595.9 & 561.5 & 19.8 & 14.7 & {\textbf{53.2}} & {\textbf{0.89}} \\ 
			& ARCS(GES) & 594 & 564 & 18 & 13 & {\textbf{49}} & {\textbf{0.89}} \\ 
			$ (300, 600, 240)$ & HC & 577 & 277 & 170 & 131 & 454 & 0.31 \\ 
			& MMHC & 455 & 307 & 143 & 5 & 298 & 0.41 \\ 
			& PC & 512 & 162 & 341 & 9 & 446 & 0.17 \\ 
			& GA & 313 & 141 & 106 & 67 & 526 & 0.18 \\ 
			\hline
	\end{tabular}
	\begin{tablenotes}[flushleft, para]
	If a method is absent for a network, that means, it took more than 10 minutes to run on a singe dataset, and thus is excluded from the comparison.
	\end{tablenotes}
	\end{threeparttable}
}
\end{table}

% ARCS(GES) vs others
\noindent{\bf ARCS(GES) versus HC, PC, MMHC and GA.}
We also compared ARCS(GES) with other existing methods, including HC, PC, MMHC and GA. Table~\ref{tab:n<p} summarizes the average performance on high-dimensional observational data. Among all the methods, ARCS(GES) achieved the lowest SHD and the highest JI for all networks.
The HC algorithm tended to output a denser DAG than the truth, leading to a large FP.
The PC algorithm had a relatively large number of reverse edges, causing a high SHD. 
The MMHC algorithm had a lower SHD than some other algorithms, but the SHD difference between ARCS(GES) and MMHC was still large. 
The PC and MMHC algorithms were slow for some networks, and thus are absent in the results for these networks.
The GA was formulated in a similar way as ARCS(GES), but the TPs of GA estimates were much lower, resulting in large SHDs for the tested networks.

Table~\ref{tab:n>p} summarizes the results for the low-dimensional setup, $n > p$. It confirms that ARCS(GES) outperforms competing algorithms by a great margin.
In general, HC had a larger P and GA had a smaller P than the true DAG, which indicates that their estimates were either too dense or too sparse. That the GA estimates were too sparse is more pronounced in this setting compared to the high-dimensional case. This might be related to its tuning parameter choice.

\begin{table}[t!]
\setlength{\tabcolsep}{4pt} 
	\centering
	\caption{ARCS against others on observational data ($n>p$).}
	\label{tab:n>p}
	\scalebox{0.9}{
	\begin{tabular}{ccrrrrrr}
		\hline
		{network} & method & P & TP & R & FP & SHD & JI \\ 
			$(p, s_0, n)$ &&&&&\\
		\hline
		\texttt{Hailfinder}   
   		% & ARCS(GES) & 300.1 & 212.8 & 40.9 & 46.5 & {\textbf{97.7}} & {\textbf{0.61}}\\ 
		& ARCS(GES) & 273 & 232 & 26 & 15 & {\textbf{47}} & {\textbf{0.76}} \\ 
 		$(224, 264, 400)$  & HC & 409 & 118 & 121 & 170 & 317 & 0.21 \\ 
   		& PC & 256 & 167 & 70 & 18 & 115 & 0.48 \\ 
   		& GA & 138 & 63 & 47 & 27 & 228 & 0.19\vspace{1mm} \\
		\texttt{Andes} 
   		% & ARCS(GES) & 342.9 & 293.9 & 24.8 & 24.1 & {\textbf{68.2}} & {\textbf{0.76}} \\
		& ARCS(GES) & 348 & 296 & 26 & 26 & {\textbf{69}} & {\textbf{0.76}} \\ 
 		$(223, 338, 400)$   & HC & 419 & 145 & 125 & 148 & 341 & 0.24 \\ 
   		& MMHC & 286 & 169 & 104 & 14 & 184 & 0.37 \\ 
   		& PC & 308 & 154 & 135 & 19 & 203 & 0.31 \\ 
   		& GA & 153 & 64 & 56 & 33 & 307 & 0.15\vspace{1mm}\\
  		\texttt{Hepar2} 
		% & ARCS(GES) & 508.1 & 339.8 & 96.8 & 71.5 & {\textbf{223.8}} & {\textbf{0.52}} \\ 
		& ARCS(GES) & 505 & 336 & 95 & 75 & {\textbf{231}} & {\textbf{0.51}} \\ 
    		(280,492,400) & HC & 535 & 152 & 180 & 203 & 544 & 0.17 \\ 
   		& PC & 354 & 125 & 180 & 49 & 416 & 0.17 \\ 
   		& MMHC & 317 & 137 & 152 & 28 & 384 & 0.20 \\
   		& GA & 416 & 114 & 84 & 218 & 597 & 0.10\vspace{1mm}\\
   		\texttt{Win95pts} 
   		% & ARCS(GES) & 473.3 & 391.6 & 45.7 & 36.0 & {\textbf{92.3}} & {\textbf{0.74}} \\ 
		& ARCS(GES) & 475 & 390 & 47 & 38 & {\textbf{96}} & {\textbf{0.73}} \\ 
  		$(304, 448, 500)$  & HC & 576 & 140 & 223 & 213 & 522 & 0.16 \\ 
   		& MMHC & 394 & 164 & 208 & 23 & 307 & 0.24 \\ 
   		& PC & 433 & 153 & 247 & 33 & 328 & 0.21 \\ 
   		& GA & 138 & 71 & 56 & 11 & 389 & 0.14\vspace{1mm} \\
  		\texttt{Pigs}   
   		% & ARCS(GES)  & 612.4 & 476.0 & 97.7 & 38.7 & {\textbf{154.7}} & {\textbf{0.66}} \\ 
		& ARCS(GES) & 619 & 471 & 102 & 46 & {\textbf{166}} & {\textbf{0.64}} \\ 
   		$(441, 592, 600)$ & HC & 843 & 377 & 164 & 302 & 517 & 0.36 \\ 
   		& GA & 218 & 112 & 83 & 23 & 503 & 0.16\vspace{1mm} \\
 		\texttt{rDAG1}   
   		% & ARCS(GES)& 301.5 & 283.5 & 15.7 & 2.4 & {\textbf{18.9}} & {\textbf{0.89}} \\ 
		& ARCS(GES) & 300 & 287 & 12 & 1 & {\textbf{14}} & {\textbf{0.92}} \\ 
   		$(300, 300, 450)$ & HC  & 549 & 186 & 105 & 257 & 371 & 0.28 \\ 
   		& MMHC & 324 & 209 & 85 & 29 & 120 & 0.51 \\ 
   		& PC & 335 & 210 & 88 & 37 & 127 & 0.50 \\ 
   		& GA & 135 & 81 & 47 & 7 & 226 & 0.20\vspace{1mm} \\
    		\texttt{rDAG2}     
   		% & ARCS(GES) & 608.3 & 580.6 & 16.2 & 11.4 & {\textbf{30.8}} & {\textbf{0.93}} \\ 
		& ARCS(GES) & 608 & 583 & 14 & 11 & {\textbf{28}} & {\textbf{0.93}} \\ 
   		$(300, 600, 600)$ & HC & 577 & 274 & 177 & 126 & 452 & 0.30 \\ 
   		& MMHC & 483 & 314 & 165 & 5 & 291 & 0.41 \\ 
   		& PC & 574 & 268 & 298 & 7 & 339 & 0.30 \\ 
   		& GA & 144 & 88 & 51 & 5 & 516 & 0.13 \\  
   		\hline
    	\end{tabular}}
	\smallskip
\end{table}

It is worth mentioning that ARCS(GES) outperformed other methods substantially for larger networks such as \texttt{Pigs} with $p = 441$ and $s_0 =592$. MMHC and PC failed to complete a single run on the \texttt{Pigs} network within 10 minutes, while HC and GA had very low accuracies. We suspect that the \texttt{Pigs} network has a certain structure that is particularly difficult to estimate, a hypothesis that merits more investigation.

We also tested another order-based algorithm, linear structural equation model learning (LISTEN) \citep{GH17}, which estimates Gaussian DAG structure by a sequential detection of ordering. A key assumption of LISTEN is that the noise variables have  equal variances. Moreover, the algorithm requires a prespecified regularization parameter for the score metric. To compare with this algorithm, we adapted our data generation process to satisfy the equal-variance assumption. ARCS(GES) achieved a much higher accuracy, with SHD as small as 14\% to 18\% of the SHD achieved by LISTEN. Full results are reported in the supplementary material.

% experimental intervention results
\subsection{Comparison on experimental data} \label{sec:expe-data}

To generate experimental datasets, we generated $p$ blocks of observations, in each of which a single variable was under intervention. For each block, we generated 5 observations, and thus $n = 5p$. 
Networks in this experiment were smaller, with $p \leq 50$ and $s_0 \leq 100$, including real DAGs $(p, s_0)$: \texttt{Asia} (8,  8),  \texttt{Sachs} (11, 17), \texttt{Ins.}(27, 52), \texttt{Alarm} (37, 46), and \texttt{Barley} (48, 84), and random DAGs: \texttt{rDAG3} (20, 20), \texttt{rDAG4} (20, 40), \texttt{rDAG5} (50, 50) and \texttt{rDAG6} (50, 100). Using these networks, we also simulated observational data of the same sample size, $n=5p$, to study the effect of experimental interventions. Since networks in the experimental setting were smaller, we used a $p$-value cutoff of $10^{-3}$ in the refinement step of the ARCS algorithm (Algorithm~\ref{algo:after-sa}).

To assess the accuracy on experimental data, we compare an estimated DAG with the true one to calculate the numbers of false positives (FP), missing edges (M), reverse edges (R) and true positives (TP). FP and M follow the same calculations as in the observational settings. R counts the number of edges whose orientations are opposite between the two DAGs, and $\text{TP} = \text{P} - \text{R}- \text{FP}$.  Note that the definition of R is different from that for observational data, because under the intervention setting used in this comparison, the true causal DAG is identifiable \citep{FZ13}. The structural Hamming distance (SHD) and the Jaccard index (JI) are then calculated as in the observational case.

In this setting, we compared ARCS with the CD algorithm \citep{FZ13} and the greedy interventional equivalence search (GIES) algorithm \citep{HB12}, both of which can handle experimental interventions. We initialize ARCS with CD and GIES estimates and call them ARCS(CD) and ARCS(GIES), respectively.

\begin{table}[t!]
\setlength{\tabcolsep}{4pt} 
	\centering
	\caption{Performance comparison on experimental data.}
	\label{tab:intv}
	\scalebox{0.9}{
	\begin{tabular}{ccrrrrrr}
  		\hline
		network & method & P & TP & R & FP & SHD & JI \\ 
		$(p, s_0, n)$  & & & & & \\
  		\hline
   		\texttt{Asia}    
   		& CD & 14 & 6 & 1 & 7 & 8 & 0.47 \\ 
 		$(8, 8, 40)$  & ARCS(CD) & 5 & 5 & 1 & 0 & {\textbf{4}} & 0.53 \\ 
   		& GIES & 9 & 1 & 6 & 2 & 9 & 0.07 \\ 
   		&ARCS(GIES) & 5 & 5 & 1 & 0 & {\textbf{4}} & {\textbf{0.53}}\vspace{1mm} \\   
    		\texttt{Sachs}   
   		& CD & 23 & 9 & 5 & 10 & 17 & 0.34 \\ 
   		$(11, 17, 55)$ & ARCS(CD) & 12 & 10 & 2 & 1 & 8 & 0.50 \\ 
   		& GIES & 17 & 4 & 10 & 4 & 17 & 0.13 \\ 
   		& ARCS(GIES) & 12 & 10 & 2 & 1 & {\textbf{8}} & {\textbf{0.51}}\vspace{1mm} \\
   		\texttt{Ins.}
   		& CD & 56 & 30 & 12 & 14 & 36 & 0.39 \\ 
   		$(27, 52, 135)$  & ARCS(CD) & 54 & 40 & 7 & 7 & {\textbf{18}} & {\textbf{0.63}} \\
   		& GIES & 72 & 14 & 34 & 24 & 62 & 0.13 \\ 
   		& ARCS(GIES) & 57 & 38 & 10 & 10 & 24 & 0.56\vspace{1mm} \\
  		\texttt{Alarm} 
   		& CD & 51 & 35 & 9 & 7 & 16 & 0.57 \\ 
   		$(37, 46, 185)$ & ARCS(CD) & 48 & 43 & 3 & 2 & {\textbf{5}} & {\textbf{0.86}} \\ 
   		& GIES & 70 & 6 & 40 & 24 & 65 & 0.05 \\ 
   		& ARCS(GIES) & 51 & 40 & 6 & 6 & 12 & 0.70\vspace{1mm} \\
    		\texttt{Barley}   
   		& CD & 85 & 52 & 17 & 16 & 48 & 0.45 \\ 
   		$(48, 84, 240)$ & ARCS(CD) & 103 & 67 & 13 & 23 & {\textbf{41}} & {\textbf{0.58}} \\ 
   		& GIES & 146 & 21 & 58 & 67 & 129 & 0.10 \\ 
   		& ARCS(GIES) & 122 & 44 & 35 & 44 & 84 & 0.28\vspace{1mm} \\
    		\texttt{rDAG3}   
   		& CD & 25 & 15 & 5 & 6 & 11 & 0.49 \\ 
   		$(20, 20,100)$ & ARCS(CD) & 20 & 18 & 1 & 1 & {\textbf{2}} & {\textbf{0.84}} \\  
   		& GIES & 29 & 4 & 15 & 10 & 25 & 0.09 \\ 
   		& ARCS(GIES) & 20 & 18 & 2 & 1 & 3 & 0.83\vspace{1mm} \\
    		\texttt{rDAG4}   
   		& CD & 43 & 25 & 8 & 10 & 24 & 0.45 \\ 
   		$(20, 40,100)$ & ARCS(CD) & 39 & 34 & 3 & 2 & 8 & 0.76 \\
   		& GIES & 48 & 6 & 30 & 11 & 46 & 0.07 \\ 
   		& ARCS(GIES) & 38 & 35 & 2 & 1 & {\textbf{6}} & {\textbf{0.82}}\vspace{1mm} \\
   		\texttt{rDAG5}   
   		& CD & 55 & 39 & 10 & 6 & 17 & 0.60 \\ 
   		$(50, 50, 250)$ & ARCS(CD) & 51 & 48 & 2 & 1 & {\textbf{3}} & {\textbf{0.90}}\\ 
   		& GIES & 87 & 7 & 43 & 37 & 80 & 0.05 \\ 
   		& ARCS(GIES) & 52 & 47 & 3 & 2 & 5 & 0.86\vspace{1mm} \\
   		\texttt{rDAG6}   
   		& CD & 101 & 70 & 17 & 13 & 43 & 0.54 \\ 
   		$(50, 100, 250)$ & ARCS(CD) & 106 & 94 & 5 & 7 & {\textbf{12}} & {\textbf{0.86}}\\ 
   		& GIES & 155 & 15 & 81 & 58 & 143 & 0.06 \\ 
   		& ARCS(GIES) & 159 & 59 & 36 & 64 & 105 & 0.34\\
   		\hline
   		\end{tabular}
	}
\end{table}

Table~\ref{tab:intv} compares the CD, ARCS(CD), GIES and ARCS(GIES) algorithms, averaging over 20 datasets for each type of networks. It is seen that both ARCS(GIES) and ARCS(CD) achieved dramatic improvements upon GIES and CD algorithms for every single network. This observation is consistent with the findings from the observational data and further confirms that the ARCS algorithm is a powerful tool for improving local estimates. Different from the observational data results (Table~\ref{tab:n<p:imp} and~\ref{tab:n>p:imp}), ARCS(CD) usually had better performance than ARCS(GIES) on experimental data. 

\begin{figure}[t]
	\begin{center}
		\includegraphics[width=.49\textwidth] {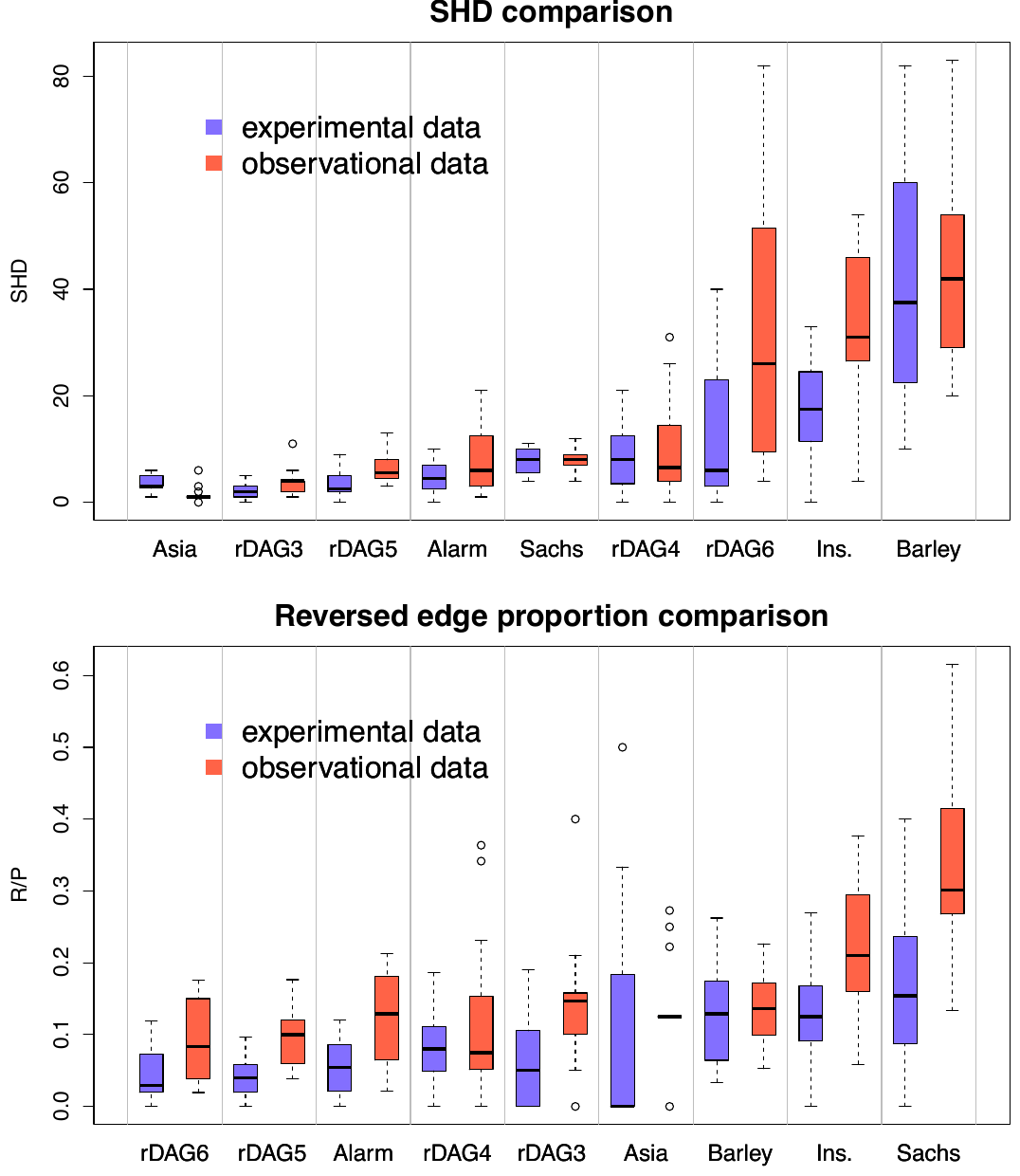}
		\caption{Comparison of SHD and reversed edge proportion between experimental and observational data with ARCS(CD). 
			% {\em Top:} SHD comparison. 
			% {\em Bottom:} Reversed edge proportion comparison.
			}
		\label{fig:intv_shd_rp}
	\end{center}
\end{figure}

\noindent{\bf Experimental versus observational.} We also compared the performance of our method ARCS(CD) on experimental and observational data with the same sample size $n=5p$. Figure~\ref{fig:intv_shd_rp} plots the SHDs of ARCS(CD) on 20 datasets, with a side-by-side comparison between observational and experimental data. For some networks, such as \texttt{rDAG6} and \texttt{Ins.}, the estimated DAGs using experimental data had much lower SHDs than using the observational data. For some small networks, such as \texttt{Asia}, ARCS(CD) achieved a low SHD on observational data and the improvement when using experimental data was not substantial.
 
Because estimated DAGs did not have the same number of predicted edges, we further compared the reversed edge proportion ($\text{R}/\text{P}$). The DAGs estimated by ARCS(CD) had a lower R/P on the experimental than the observational data for all networks. The decrease in R/P with experimental interventions was remarkable, as Figure~\ref{fig:intv_shd_rp} shows. This finding supports the idea that experimental interventions help correct the reversed edges and distinguish equivalent DAGs. Note that for the 20 observational datasets generated from \texttt{Asia} ($p=8$, $s_0 = 8$), ARCS(CD) output 16 estimated DAGs with $\text{P}=8$ and $\text{R}=1$, resulting in a very thin interquantile range in the boxplots of \texttt{Asia} in Figure~\ref{fig:intv_shd_rp}.

% BIC
\subsection{Effectiveness of BIC selection}

Given an initial permutation, we use the BIC to choose tuning parameters $(\gamma, \lambda)$ before applying the ARCS algorithm (Section \ref{sec:bic}). In Tables~\ref{tab:n<p} and~\ref{tab:n>p}, the number of predicted edges by ARCS(GES) is closer to $s_0$ than any other competing method in every network. This observation signifies the effectiveness of our parameter selection method by BIC. To further study its effect, we compared DAGs estimated with all values on a grid of $(\gamma, \lambda)$ by ARCS(CD) and ARCS(GES).

\begin{figure}%[t]
	\centering
	\includegraphics[width=.35\textwidth] {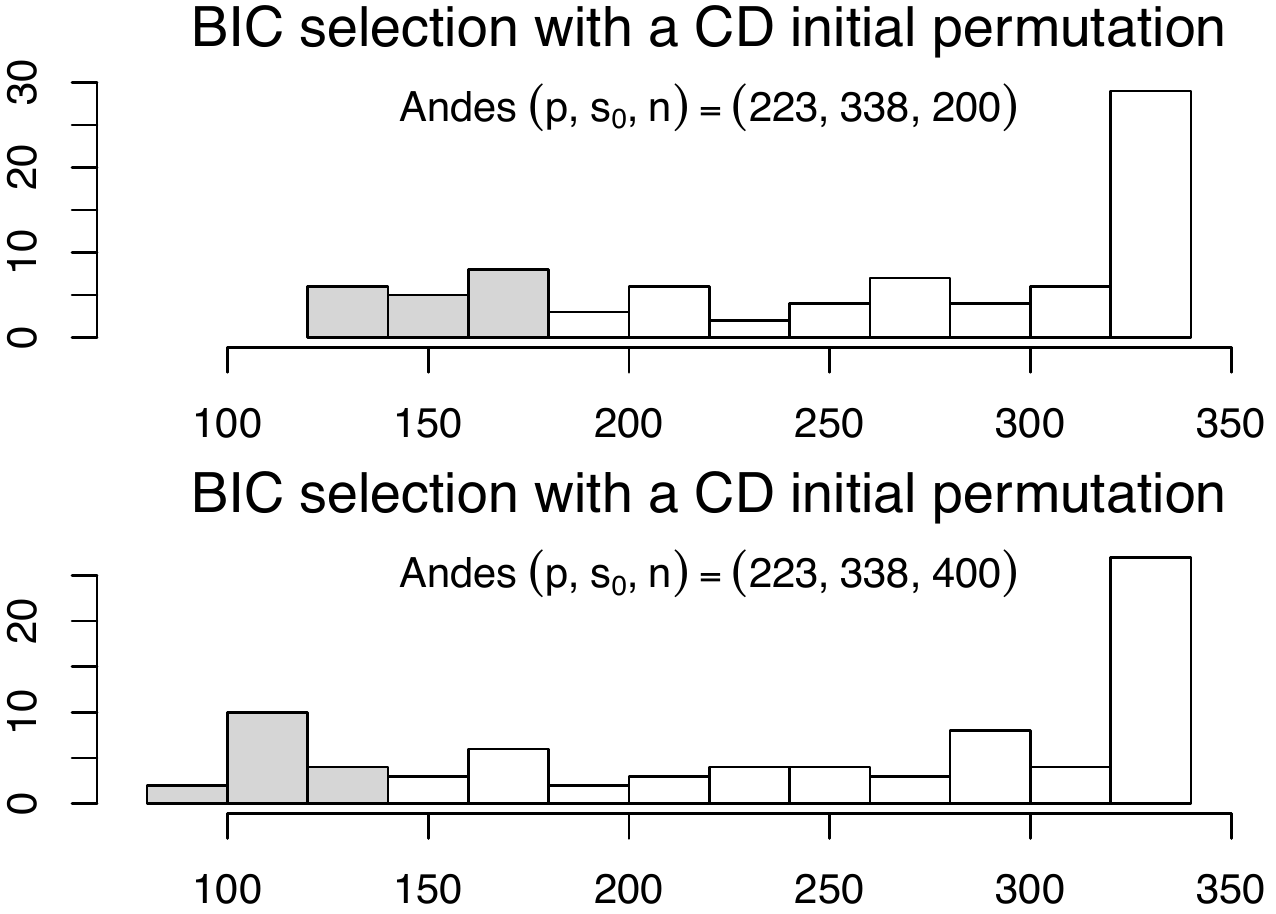}\\
	\vspace{2mm}
	\includegraphics[width=.35\textwidth] {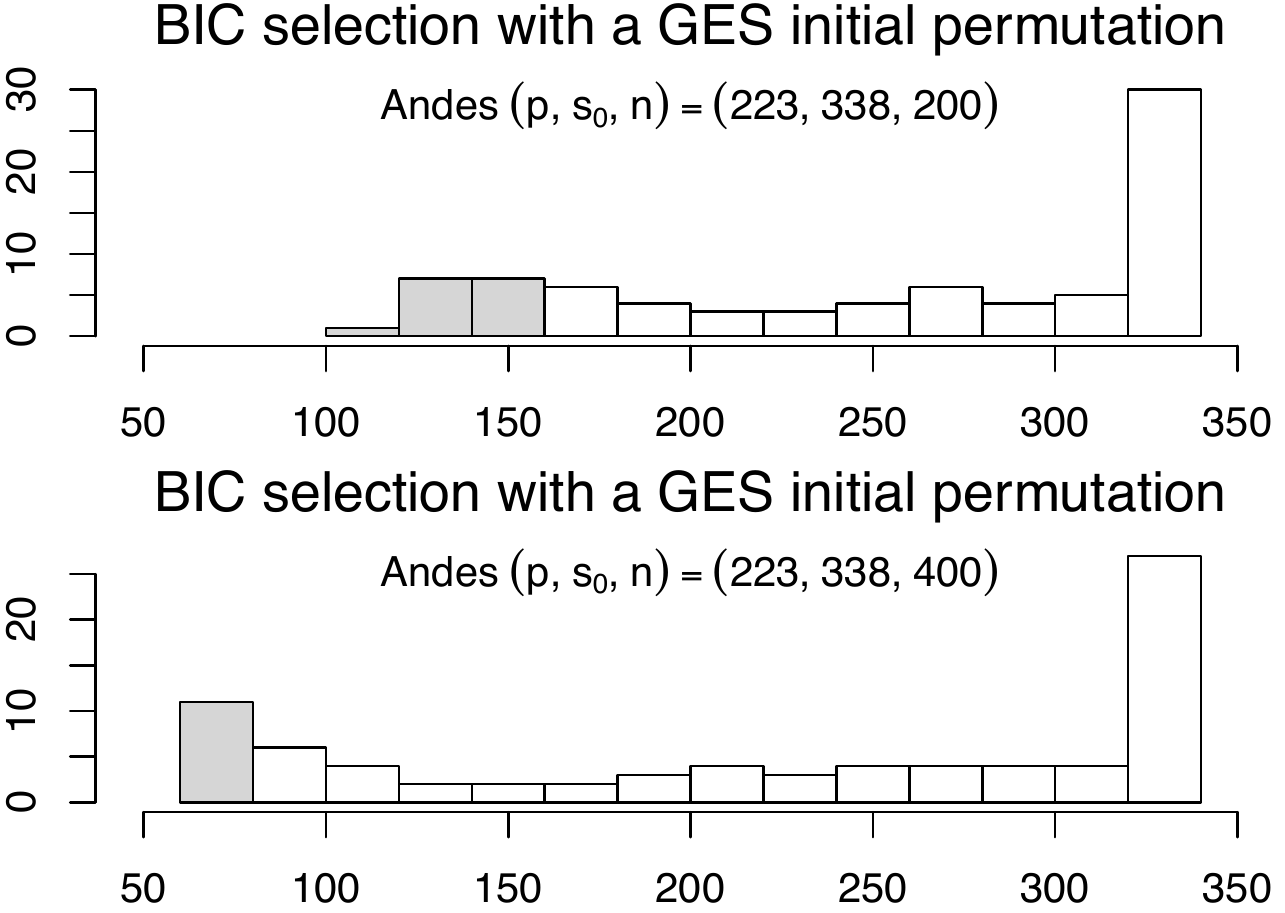}
	\caption{Performance of the BIC selection among a grid of $(\gamma, \lambda)$ given an initial permutation. Tuning parameters that lead to lower SHDs than the BIC selection are shown in gray.}
	\label{fig:grid_par}
\end{figure}

Figure~\ref{fig:grid_par} shows the histograms of the SHDs for all values of $(\gamma, \lambda)$ on \texttt{Andes} datasets with $(p,s_0) = (223, 338)$ and $n \in \{200, 400\}$. The grey area in a histogram corresponds to values of $(\gamma, \lambda)$ that led to a lower SHD than the $(\gamma^*, \lambda^*)$ chosen by the BIC, i.e., it indicates the percentile of the SHD of the BIC selection among all choices of the tuning parameters. The smaller this percentile, the better the BIC selection performance. 

Each histogram has a high spike of large SHD, corresponding to large values of $\lambda$ that generate almost empty graphs. The SHDs of ARCS(CD) with BIC selection corresponded to the $14^\text{th}$ and $16^\text{th}$ percentiles in low and high dimensions, respectively, and for ARCS(GES) the $16^\text{th}$ and $3^\text{nd}$ percentiles. These low percentiles confirm that the BIC selection works well for choosing the tuning parameters. 
Moreover, in our tests, the BIC usually selected $\gamma^*=2$, the smallest provided value for $\gamma$. Since for small $\gamma$, the MCP is closer to the $\ell_0$ penalty and far from the $\ell_1$ norm, this choice of $\gamma$ indicates the preference of concave penalties over $\ell_1$ in estimating sparse DAGs. Some of CD's and GA's inferior performances, such as CD on the \texttt{Pigs} network (Table~\ref{tab:n<p:imp}) and the overall performance of GA (Tables~\ref{tab:n<p} and \ref{tab:n>p}), were potentially caused by a bad choice of their tuning parameters. This demonstrates the importance of our data-driven selection scheme for a regularized likelihood method.

% RND
\subsection{Random initialization with a high temperature}
Recall that we started ARCS(CD) and ARCS(GIES) with $T^{(0)} = 1$. To test its global search ability over the permutation space, we may initialize the annealing process with a random permutation and a high temperature, which we denote by ARCS(RND). For a random initial permutation, we do not need to preserve its properties, so we use a high initial temperature $T^{(0)} = 100$ to help the algorithm traverse the search space. 

\begin{figure}[t]
	\begin{center}
		\includegraphics[width=.49\textwidth] {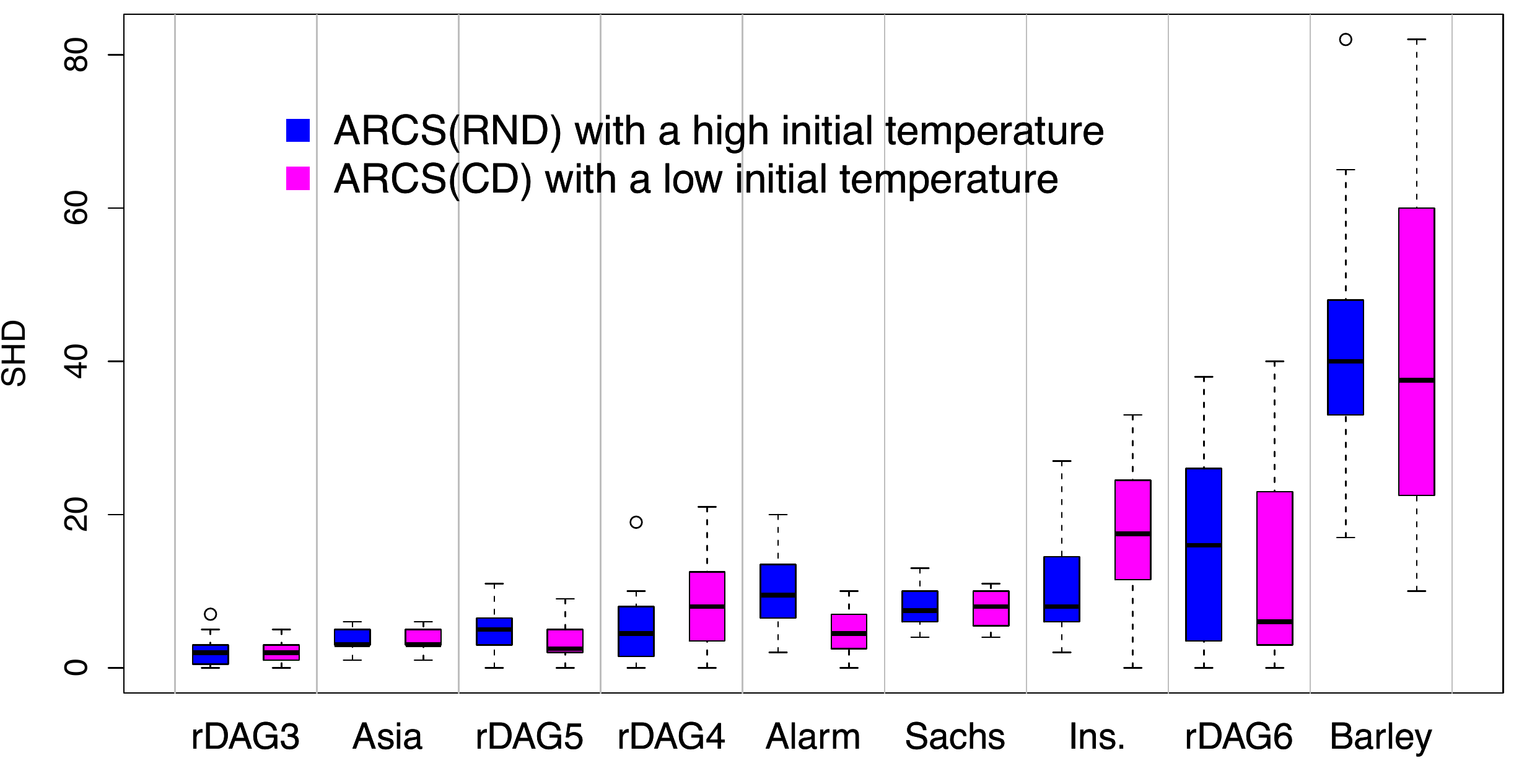}
		\caption{A comparison between ARCS(RND) with a high initial temperature and ARCS(CD) on experimental data.}
		\label{fig:hight}
	\end{center}
\end{figure}

We compared ARCS(RND) and ARCS(CD) on experimental data for nine networks. We chose  ARCS(CD) due to its superior performance on these networks in the experimental setting (Table~\ref{tab:intv}). As shown in Figure \ref{fig:hight}, ARCS(RND) and ARCS(CD) had comparable performances on all networks. Both of them learned DAGs with small SHDs. ARCS(RND) was slightly better on \texttt{rDAG4} and \texttt{Ins.}, and slightly worse on \texttt{Alarm} and \texttt{rDAG6}. For the other networks, the two methods were quite comparable, demonstrating the effectiveness of ARCS(RND).

The networks in this experiment had $p \leq 50$. For $p = 50$, there are $50! \approx 3 \times 10^{64}$ possible permutations. ARCS(RND) managed to learn a network structure within $10^4$ iterations, which is much smaller than $p!$. 
However, the performance of ARCS(RND) was not competitive on large networks. The reason is that for large $p$, it takes much more time for the annealing to thoroughly search the huge permutation space. Therefore, for large networks, it is better to initialize the ARCS algorithm with estimates from other local methods and choose a low temperature. 
Given a good initial estimate, ARCS searches over the permutation space and improves the accuracy of the initial estimate as demonstrated in Tables~\ref{tab:n<p:imp},~\ref{tab:n>p:imp} and~\ref{tab:intv}. This study suggests that, by combining effective local and global searches under a regularized likelihood framework, our ARCS algorithm is a promising approach to the challenging problem of DAG structure learning.

\section{Discussion}\label{sec:discussion}

In this paper, we developed a method to learn Gaussian BN structures by minimizing the MCP regularized Cholesky score over topological sorts, through a joint iterative update on a permutation matrix and a lower triangular matrix. We search over the permutation space and optimize the network structure encoded by a lower triangular matrix given a topological sort. This approach relates BN learning problem to sparse Cholesky factorization, and provides an alternative formulation for the order-based search. 
Our method can serve as an improvement of a local search or a stand-alone method with a best-guess initial permutation. Although we formulated this order-based search for Gaussian BNs, it can potentially be extended to discrete BNs and other scoring functions. A main difference in the extension to discrete data is the proximal gradient step, where we can borrow the regularized multi-logit model in \cite{GFZ18} or develop a continuous regularizer for multinomial likelihood.

For the simulated annealing step, there are several interesting aspects to investigate. For instance, various operators of moving from one permutation to another have been proposed for greedy order-based search, which could better guide the annealing process in traversing the search space. 
This paper focuses on numerical results, and shows the advantage and potential application of local search and global annealing in learning BNs. Left as future work are theoretical properties of our method, such as the consistency of the regularized Cholesky score and the convergence of ARCS with a good initial permutation.

%%%%%%%%%%%%%%%%%%%%%%%%%%%%%%%%%%%%%%%

% or


\appendices

\section{}\label{app:proofs}

%% proof of the permutation invariance
\subsection{Proof of Proposition~\ref{prop:cf}}\label{proof:prop:cf}

Since $A$ is positive definite, $A^{-1}$ is positive definite as well. Let us apply the Cholesky decomposition on $A^{-1}$, i.e., $A^{-1} = CC^\top$ where $C= \calC(A^{-1})$. Therefore, $A = C^{-\top}C^{-1}$. Recall that $\calL_p$ is the set of lower triangular matrices with positive diagonals. The goal is to show that the minimizer of $\lchol(L; A)$ over $\calL_p$, denoted by $L^*$, equals to $C$.

Letting  $R = C^{-1}L$, we have $\log |R| = -\log |C| + \log|L|$, where $\log |C|$ is a constant. 
Since $C \in \calL_p$ and $L \in \calL_p$, we have $R \in \calL_p$ as well. Also,
 \begin{align*}
\tr( A L L^\top) & = \tr(C^{-\top} C^{-1}\ L L^\top )= \tr(RR^T)=  \norm{R}_F^2.
\end{align*}
 Then, 
 \begin{align*}
	& \min_{L \in \calL_p} \lchol(L;A) = \min_{L \in \calL_p} \left[\f 1 2 \tr( A L L^\top)  -\log |L| \right] \\
	& = \min_{R \in \calL_p} \f 1 2 \Big[ \norm{R}_F^2 - 2 \log|R| \Big]  - \log |C| \\
	& = \min_{R \in \calL_p} \f 1 2 \Big[ \sum_{(i,j): i>j} R_{ij}^2 + \summ i p \left( R_{ii}^2 - 2 \log R_{ii}\right)\Big] -\log |C|.
%	& {~~~~~} - \log |C|.
\end{align*}

The problem is separable over entries. 
Minimizing over off-diagonal entries, the unique minimizer is $R_{ij}^* = 0$ for all $i > j$ (and clearly $i < j$). 
Optimizing over $R_{ii}$, we are minimizing $x \mapsto x^2 - 2 \log x$ over $x > 0$. 
The unique minimizer is attained at $x = 1$, i.e. $R_{ii}^* = 1$ for all $i$. 
Therefore the unique minimizer is $R^* = I_p$, i.e.  $C^{-1} L^*= I_p$. 
In terms of the original variable, the minimizer is $L^* = C =  \calC(A^{-1})$.
	
Substituting  $L^* = \calC(A^{-1})$ into $\lchol(L;A)$, we obtain the minimum value:
\begin{align*}
	\lchol^*(A)& := \lchol(L^*;A)
		=  \f 1 2\tr\left( AA^{-1}\right)  -\log |A^{-\f 1 2}| \\
		& =\f 1 2 \left(p + \log|A|\right).
\end{align*}
The assertion $\lchol^*(A) = \lchol^*(PAP^\top)$ follows by noting that $|P \Sigh P^T| = |P| |\Sigh| |P^\top| = |\Sigh|$.

%% derivation of intv likelihood
\subsection{Proof of Lemma~\ref{lem:interv:loglike}} \label{proof:exp:like}

Recall $B = (\beta_j) \in \R^{p\times p}$, $\Omega = \diag(\omega_j^2) \in \R^{p\times p}$, we have the following experimental data log-likelihood:
\begin{align*}
	 &{~} \ell_{\calO}(B,\Omega, P_\pi \mid \Xb) \\
		= &{~} \sum_{j=1}^p \frac1{2\omega_{j}^2} \norm{\Xb_{\calO_{\pi(j)}, \pi(j)}  - \Xb_{\calO_{\pi(j)}} P_\pi^\top \beta_j}^2  + \f 1 2 |\calO_{\pi(j)}| \log \omega_{j}^2 \\
		 = &{~} \f 1 2 \summ j p  \bigg[  \frac1{\omega_{j}^2} \norm{\Xb_{\calO_{\pi(j)}}P_\pi^\top (e_j - \beta_j)}^2 + |\calO_{\pi(j)}|  \log \omega_{j}^2 \bigg] \\
		 = &{~} \f 1 2 \summ j p  \bigg[ \frac1{\omega_{j}^2} \tr\left( P_\pi \Xb_{\calO_{\pi(j)}}^\top \Xb_{\calO_{\pi(j)}} P_\pi^\top (e_j - \beta_j) (e_j - \beta_j)^\top\right) \\
		&{~~~~~~~~~~~~~~~~} +  |\calO_{\pi(j)}| \log \omega_{j}^2 \bigg].\\
\end{align*}
Recall $L_j = (e_j - \beta_j)/\omega_j \in \R^p$ and $L = (L_j)\in \R^{p\times p}$. Define $|L_j| := L_{jj}$. 
Recall $\Sigh^{j} := \frac1{|\calO_{\pi(j)}|} \Xb_{\calO_{\pi(j)}}^\top \Xb_{\calO_{\pi(j)}}$ and simplify $P_\pi$ as $P$. We have
\begin{align*}
	\ell_{\calO}(L, P) & = \f 1 2 \summ j p  |\calO_{\pi(j)} | \cdot \left[  \tr\big( P\Sigh^{j}P^\top  L_j L_j^\top\big) -\log L_{jj}^2 \right] \\
		& = \summ j p  |\calO_{\pi(j)}|  \left[ \f {1}  2\tr\left(  P\Sigh^{j}P^\top L_j L_j^\top\right)  -  \log |L_j| \right] \\
		& = \summ j p   |\calO_{\pi(j)}| \lchol\left(L_j; P\Sigh^{j}P^\top \right),
\end{align*}
where $\lchol(L; A)$ is the Cholesky loss defined in~\eqref{eq:chol:loss}.

%% proof of gradients 
\subsection{Proof of Lemma~\ref{lemma:gradl}} \label{proof:lemma:gradl}

Fix $P$ and let $Z = \f 1 {\sqrt{n}} \Xb P^\top$ so that $Z^\top Z = \f 1 n P \Xb^\top \Xb P^\top = P \widehat\Sigma P^\top$. 
We can rewrite~\eqref{eq:LP:loss} as
\begin{align*}
	\ell( L) := \ell( L, P) = \f n 2 \norm{Z L}_F^2  -n\log |L|.
\end{align*}
Let $\ell^1(L) :=  \f n 2 \norm{Z L}_F^2$ and $\ell^2(L) := -n\log |L|$. 
To compute the gradient of $\ell^1$ w.r.t. $L$, we perturb $L$ to $L + t \delta$ where $\delta\in\calL_p$ and $t\in\R$. We have 
\begin{align*}
\lim_{t \to 0}\f{ \ell^1(L+t\delta) - \ell^1(L)}{t} & = \lim_{t \to 0} \Big[{ n} \left\langle Z L, Z \delta \right\rangle +  \f {t n} 2 \norm{Z\delta}_F^2\Big]  \\
&=  n  \big\langle ZL, Z \delta \big\rangle = n \big\langle Z^\top Z L, \delta \big\rangle,
\end{align*}
where $\langle A, B \rangle = \tr(A^\top B)$ for two matrices $A$ and $B$. 
% Be definition, $\Pi_\calL: A \mapsto (A_{ij} \mathbbm{1}\{i \geq j\})_{p\times p}$ maps a $p \times p$ matrix to its lower triangular projection. 
Be definition, $\Pi_\calL$ maps a $p \times p$ matrix to its lower triangular projection. 
Since $\delta \in\calL_p$, $\langle Z^\top Z L, \delta\rangle = \langle \Pi_\calL (Z^\top Z L), \delta \rangle$, and thus $\nabla \ell^1 (L) = n  \Pi_\calL(Z^\top Z L) $. 

For the second term, $\ell^2 = -n \summ i p \log L_{ii}$, and thus $ \nabla \ell^2 (L) = -n \diag\left(\{1/L_{ii}\}_{i=1}^p\right)$ where $L\in\calL_p$. Putting the pieces together gives the desired result.

\subsection{Proof of Lemma~\ref{lemma:intv_grad}}  \label{proof:intv_grad}

From~\eqref{eq:LP:ellO}, we have $\ell_\calO =  \ell_\calO^1 +  \ell_\calO^2$, where
\begin{align*}
	 \ell_\calO^1 &:=  \f 1 2 \sum_{j=1}^p  |\calO_{\pi(j)}|  L_j^\top P \widehat\Sigma^j P^\top L_j , \\
 	%	& \quad \text{and} \quad \\
	 \ell_\calO^2 &:=  - \sum_{j=1}^p  |\calO_{\pi(j)}|  \log L_{jj}.
\end{align*}
Note that each term in $\ell_\calO^1$ is a quadratic form in $L_j$.
The gradient of $\ell^1_\calO$ w.r.t. to $L_j$ is $ |\calO_{\pi(j)}|  \Pi_j (P \widehat\Sigma^j P^\top L_j)$, since $L_{ij} = 0$ for $i<j$.
It is also easy to see that  $\nabla_{L_j}\ell^2_\calO =- |\calO_{\pi(j)}| \f {e_j} {L_{jj}}$. 
Putting the pieces together, the gradient of $\ell_\calO$ w.r.t. $L_j$ is 
$|\calO_{\pi(j)}|  \left(\Pi_j(P\widehat \Sigma^j P^\top L_j) - \frac {e_j} {L_{jj}}\right)$.

%% proof of prox-mcp 
\subsection{Proof of Lemma~\ref{lemma:prox_mcp}} \label{app:prox_mcp}

Recall the MCP function is defined as
\[ 
	\rho(x) =  
		\begin{cases} 
			\lambda |x| - \f {x^2} {2\gamma}, & \mbox{$|x| < \gamma\lambda$,}\\
			\f 1 2 \gamma\lambda^2, &\mbox{$|x|\geq \gamma\lambda$.}
		\end{cases}
\]
It is not hard to see that $\rho(\lambda \gamma x) = \gamma \lambda^2  \rho_1(x)$.
It follows that
\begin{align*}
	& {~~~~} \prox_{t \rho}( \lambda \gamma x) \\
	&= \argmin_{u} \Big[ \rho(u) + \frac1{2t} (u- \lambda \gamma x)^2 \Big] \\
	&= \lambda \gamma \cdot \argmin_{ v} \Big[ \rho(\lambda \gamma v) + \frac1{2t} (\lambda \gamma v- \lambda \gamma x)^2 \Big] \\
	&= \lambda \gamma \cdot \argmin_{ v} \Big[ \gamma\lambda^2  \rho_1( v) + \frac{\gamma^2\lambda^2 }{2t} ( v- x)^2 \Big]\\
	&= \lambda \gamma \cdot \argmin_{ v} \Big[ \rho_1( v) + \frac{\gamma}{2t} ( v- x)^2 \Big] \\
	&= \lambda \gamma \prox_{(t/\gamma) \rho_1}(x),
\end{align*}
where the second equality uses the change of variable $u = \lambda \gamma v$ and the fourth uses the fact that $\argmin$ is invariant to rescaling the objective. This establishes~\eqref{eq:prox-lemma}.

Due to the symmetry of $\rho_1$, we have $\prox_{\alpha \rho_1}(-x) = - \prox_{\alpha \rho_1}(x)$, which is easy to see by a change of variable $u = -v$ in the defining optimization. Thus, it is enough to consider $x \geq 0$ which we assume in the following.

\begin{figure}[t]
	\centering
	\begin{tabular}{ccc}
		\includegraphics[width=.485\textwidth]{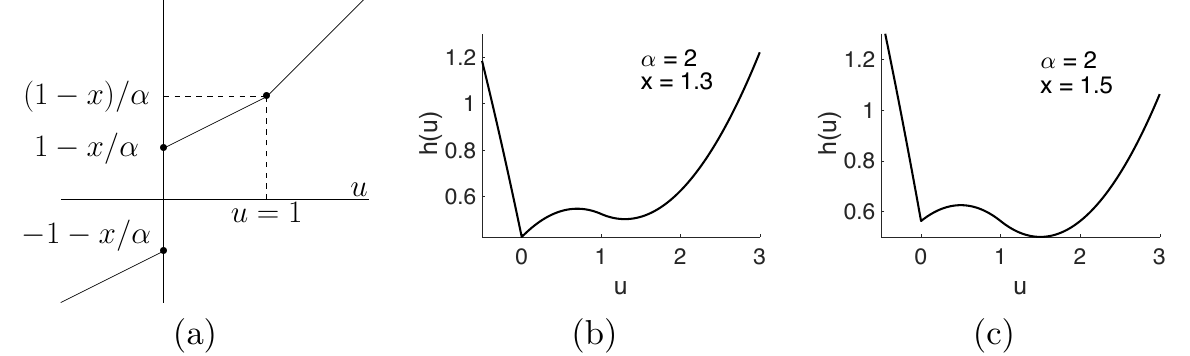} 
		\end{tabular}
	\caption{(a) The derivative $h'(u)$ for $u \in (0,\infty)$ in a typical case. Note that $h'$ is discontinuous at $0$. (b) and (c) Plots of $h(u)$ for two values of $(\alpha,x)$ that lead to case 4. }
	\label{fig:h:prime}
\end{figure}

The function $h(u) = \rho_1(u) + \frac1{2\alpha} (u-x)^2$ is continuous and decreasing on $(-\infty,0]$, hence it achieves its minimum of $\frac{x^2}{2\alpha}$ over this interval at $u = 0$. Over $(0,\infty)$, the function is differentiable with derivative $h'(u) = (1-u)_+ + \frac{1}{\alpha}(u-x)$ which is piecewise linear (or affine) with a break at $u=1$. The first segment of $h'$ is a line connecting $(0,1-x/\alpha)$ to $(1,(1-x)/\alpha)$. The next segment is an always-increasing section starting at  $(1,(1-x)/\alpha)$ and increasing with slope $1/\alpha$. See Figure~\ref{fig:h:prime}(a). The behavior of $h'$ for $u \in (0,1)$ determines the minimizer. We have four cases:
\begin{enumerate}
	\item When both $1-x/\alpha > 0$ and $(1-x)/\alpha > 0$, that is $x < \min\{\alpha,1\}$, $h'$ is increasing in $[0,1]$. Hence, $h' > 0$ over $[0,\infty)$ and the overall minimum of $h$ occurs at $u_1 = 0$. 
	\item When $1-x/\alpha \leq 0 < (1-x)/\alpha$, that is, $\alpha \leq x < 1$, then $h$ has a single critical point at $u_2 = (x-\alpha)/(1-\alpha)$ before which it decreases and after which it increases. Hence, this is its unique minimizer.
	\item When both $1-x/\alpha < 0$ and $(1-x)/\alpha < 0$, that is, $x > \max\{\alpha,1\}$, then, the only critical point occurs in the second linear segment and is $u_3 =x$ which is the unique minimizer.
	\item When $(1-x)/\alpha < 0 \leq 1-x/\alpha$, that is, $1 < x \leq \alpha$, then both of the points $u_2$ and $u_3$ in cases 2 and 3 are critical points. The function drops in $(-\infty,u_1)$ where $u_1=0$, increases in $(u_1,u_2)$, drops in $(u_2,u_3)$ and increases in $(u_3,\infty)$. Thus, both $u_1$ and $u_3$ are local minima (while $u_2$ is a local maximum). The global minimum is determined by comparing $h(u_1) = x^2/(2\alpha)$ and $h(u_3) = 1/2$. That is, if $ x < \sqrt{\alpha}$, the global minimum is $u_1 = 0$ (Figure~\ref{fig:h:prime}b);  if $ x > \sqrt \alpha$, the global minimum is $u_3 = x$ (Figure~\ref{fig:h:prime}c); if $x = \sqrt{\alpha}$, minimum is $\{u_1,u_3\} = \{0,x\}$, which is not unique. 
\end{enumerate}
What left is the special case when both $1-x/\alpha = 0$ and $(1-x)/\alpha = 0$, i.e. $\alpha = x = 1$, indicating the first segment of $h'(u)$, $u\geq0$, is flat as zero. Hence, minimizer of $h$ is any value in the interval $[0,1]$.
We merge case 4 into cases 1 and 3 by revising the domains of $x$, and thus complete the derivation of \eqref{eq:prox-mcp-formula}.

% \section{}

\iffalse
\ifCLASSOPTIONcompsoc
  \section*{Acknowledgments}
\else
  \section*{Acknowledgment}
\fi

The authors would like to thank...
\fi

\ifCLASSOPTIONcaptionsoff
  \newpage
\fi

\bibliographystyle{IEEEtran}
\bibliography{./cite2}

\end{document}

%% file: defs.tex
\newcommand{\given}{\mid}
\newcommand\Xb{\mathbf{X}}
\newcommand\Lc{\mathscr L}
\newcommand\lchol{\Lc_{\text{chol}}}
\newcommand\Sigh{\widehat \Sigma}
\newcommand\Ph{\widehat P}
\newcommand\pih{\widehat \pi}

\newcommand{\f}{\frac}

\newcommand{\R}{\mathbb{R}}
\newcommand{\eps}{\varepsilon}
\newcommand{\norm}[1]{\left\|#1\right\|}

\newcommand{\summ}[2]{\sum_{#1 = 1}^{#2}}

\newcommand{\diag}{\operatorname{diag}}

\newcommand{\calC}{\mathcal{C}}

\newcommand{\calG}{\mathcal{G}}

\newcommand{\calI}{\mathcal{I}}

\newcommand{\calL}{\mathcal{L}}

\newcommand{\calN}{\mathcal{N}}
\newcommand{\calO}{\mathcal{O}}
\newcommand{\calP}{\mathcal{P}}

\usepackage{array}
\newcolumntype{H}{>{\setbox0=\hbox\bgroup}c<{\egroup}@{}} % hide a colum

\newtheorem{lemma}{Lemma}
\newtheorem{prop}{Proposition}
\newtheorem{remark}{Remark}

\DeclareMathOperator{\tr}{tr}

\DeclareMathOperator*{\argmin}{arg\,min}
\DeclareMathOperator{\ex}{\mathbb E}
\DeclareMathOperator{\prox}{\mathbf{prox}}

\usepackage{xspace} 
\newcommand{\MATLAB}{\textsc{Matlab}\xspace}